\documentclass[11pt,a4paper]{article}
\usepackage{times}
\usepackage{latexsym}
\usepackage{url}
\usepackage{placeins}
\usepackage{csquotes}

\usepackage{mathtools}
\usepackage{setspace}
\usepackage{dsfont}
\usepackage{amsfonts}
\usepackage{amsmath}
\usepackage{subcaption}
\usepackage{paralist}
\usepackage{times}
\usepackage{latexsym}
\usepackage{graphicx}
\usepackage[T1]{fontenc}
\usepackage{tikz}
\usepackage{tikz-dependency}

\usepackage{url}
\usepackage{pgfplotstable}
\usepackage{color}
\usepackage{lipsum,adjustbox}
\usepackage[font={small}]{caption}
\usetikzlibrary{positioning}
\usepackage{bbm}
\usepackage{multirow}
\usepackage{pdflscape}
\usepackage{xr}
\makeatletter
\newcommand*{\addFileDependency}[1]{
  \typeout{(#1)}
  \@addtofilelist{#1}
  \IfFileExists{#1}{}{\typeout{No file #1.}}
}
\newcommand*{\myexternaldocument}[1]{%
    \externaldocument{#1}%
    \addFileDependency{#1.tex}%
    \addFileDependency{#1.aux}%
}
\myexternaldocument{appendix}
\makeatother

\makeatletter
\newcommand{\@BIBLABEL}{\@emptybiblabel}
\newcommand{\@emptybiblabel}[1]{}

\usepackage[compact]{titlesec}
\titlespacing{\section}{0pt}{0x}{0ex}
\titlespacing{\subsection}{0pt}{0ex}{0ex}

\usepackage[draft]{hyperref}
\usepackage[hyperref]{acl2020}
\graphicspath{{./plots/}}
\newcommand{\com}[1]{}

\newcommand{\ttt}[1]{{\texttt{#1}}}

\newenvironment{myequation*}{
	\vspace{-1em}
	\begin{equation*}
}{
\end{equation*}
\vspace{-1.2em}
}

\aclfinalcopy

\begin{document}

\title{Fine-Grained Analysis of Cross-Linguistic Syntactic Divergences}
\newcommand{\emldisplay}[2]{\texttt{\href{mailto:#1}{#2}}}

\newcommand{\affa}{{$^{\dagger}$}}
\newcommand{\affb}{{$^{\ddagger}$}}

\author{
Dmitry Nikolaev\affb\thanks{~~~Work mostly done while at the Hebrew University of Jerusalem.} \quad\quad Ofir Arviv\affa \quad\quad Taelin Karidi\affa  \quad\quad {\bf Neta Kenneth}\affa \\  {\bf Veronika Mitnik}\affa \quad\quad {\bf Lilja Maria Saeboe}\affa \quad\quad {\bf Omri Abend}\affa \\
  \affb Stockholm University \quad\quad
  \affa Hebrew University of Jerusalem \\
  \texttt{dnikolaev@fastmail.com}\\
  \texttt{\{ofir.arviv|taelin.karidi|neta.kenneth|}\\
  \texttt{veronika.mitnik|liljama.saeboe|omri.abend\}@mail.huji.ac.il}\\
}

\maketitle

\begin{abstract}
        The patterns in which the syntax of different languages converges and diverges are often used to inform work on cross-lingual transfer.
        Nevertheless, little empirical work has been done on quantifying the prevalence of different syntactic divergences across language pairs.
        We propose a framework for extracting divergence patterns for any language pair from a parallel corpus, building on Universal Dependencies \citep[UD;][]{nivre2016universal}. We show that our framework provides a detailed picture of cross-language divergences, generalizes previous approaches, and lends itself to full automation.
        We further present a novel dataset, a manually word-aligned subset of the Parallel UD corpus in five languages, and use it to perform a detailed corpus study. We demonstrate the usefulness of the resulting analysis by showing that it can help account for performance patterns of a cross-lingual parser.
\end{abstract}

\section{Introduction}

The assumption that the syntactic structure of a sentence is predictably related to the syntactic structure of its translation has deep roots in NLP, notably in cross-lingual transfer methods, such as annotation projection and multi-lingual parsing \citep[\textit{inter alia}]{Hwa:05,mcdonald2011multi,kozhevnikov2013cross,rasooli2017cross}, as well as in syntax-aware machine translation \citep[MT;][]{birch2008predicting,williams2016syntax,bastings2017graph}. Relatedly, typological parameters that provide information on the dimensions of similarity between grammars of different languages were found useful for a variety of NLP applications \citep{ponti2018modeling}. For example, neural MT in low-resource settings has been shown to benefit from bridging morphosyntactic differences in parallel training data by different types of preprocessing, such as reordering \citep{zhou-etal-2019-handling} and hand-coded syntactic manipulations \citep{ponti2018isomorphic}.

Nevertheless, little empirical work has been done on systematically quantifying the type and prevalence of syntactic divergences across languages. Moreover, previous work generally classified divergences into a small set of divergence classes, often based on theoretical considerations \citep{dorr1994machine} or on categorical (``hard'') typological features selected in an ad-hoc manner, and left basic questions, such as how often POS tags are preserved in translation and what syntactic structures are likely correspondents of different syntactic relations, largely unaddressed. See \S~\ref{sec:related_work}.

\begin{figure}
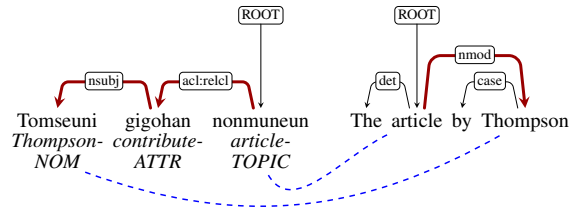

\scalebox{.7}{
\begin{dependency}
\begin{deptext}
Tomseuni \& gigohan \& nonmuneun \&[.6cm] The \& article \& by \& Thompson \\
{\it Thompson-} \& {\it contribute-} \& {\it article-} \\
{\it NOM} \& {\it ATTR} \& {\it TOPIC} \\
\end{deptext}
\depedge[edge style={red!60!black,ultra thick}]{2}{1}{nsubj}
\depedge[edge style={red!60!black,ultra thick}]{3}{2}{acl:relcl}
\deproot{3}{ROOT}
\deproot{5}{ROOT}
\depedge{5}{4}{det}
\depedge[edge style={red!60!black,ultra thick}]{5}{7}{nmod}
\depedge{7}{6}{case}
\draw[thick, dashed, blue] (\wordref{3}{1})  to[out=-20, in=-150] (\wordref{1}{7});
\draw[thick, dashed, blue] (\wordref{3}{3})  to[out=-60, in=-155] (\wordref{1}{5});
\end{dependency}
}
\vspace{-1.0cm}\caption{An example En-Ko sentence pair exhibiting a divergence, where an En \ttt{nmod} path corresponds to a Ko \ttt{acl:relcl}+\ttt{nsubj} path (paths are given in bold red). The En preposition {\it by} is not considered a content word and is not aligned with the Ko verb {\it gigohan}. See \S~\ref{sec:clmd} for further details. \label{fig:divergence_ex} \vspace{-.5cm}}
\end{figure}

We propose a language-neutral, fine-grained definition of \textit{cross-linguistic morphosyntactic divergences} (CLMD) that allows for their extraction using a syntactically annotated, content-word-aligned parallel corpus.
Concretely, we classify CLMD based on the edge labels on the dependency paths between corresponding pairs of content words (\S~\ref{ssec:defining_divergences}). See Figure~\ref{fig:divergence_ex} for an example.\footnote{It may appear that divergences recoverable by means of UD edge labels are purely \textit{syntactic} and not morphosyntactic. However, this is not the case: the domain of \enquote{pure syntax} is not well defined in a non-theoretical perspective, and many phenomena we are dealing with, e.g.\ a switch from a direct-object construction to an oblique construction, often involve morphological processes, such as adding a case ending.}

We further conduct a detailed corpus study, manually aligning content words in a subset of the PUD corpus \citep{conll2017shared} over five language pairs---English-French (En-Fr), English-Russian (En-Ru), English-Chinese (En-Zh), English-Korean (En-Ko), and English-Japanese (En-Jp) (\S~\ref{ssec:aligned_pud})---and analyze the prevalence of divergences by types (\S~\ref{sec:corpus_study}). The resulting resource can be useful for MT research, by guiding the creation of challenge sets focusing on particular constructions known to be cross-linguistically divergent, as well as by guiding preprocessing of source-side sentences for better MT performance.\footnote{The resource can be found at \url{https://github.com/macleginn/exploring-clmd-divergences}}

The emerging CLMD provide information not only on the macro-structure of the grammar (e.g., whether the language is pro-dropping), but also on parameters specific to certain lexical classes (e.g., modal verbs) and probabilistic tendencies (e.g., Japanese tends to translate sequences of events, expressed in English using coordinating conjunctions, with subordinate clauses). See \S~\ref{sec:refining-ud}.

Further experiments demonstrate the methodology's applicative potential.
First, we show that the proposed methodology can be straightforwardly automated
by replacing manual parses and alignments with automatically induced ones
(\S~\ref{sec:automation}). We present a study done on a larger En-Zh corpus,
which yields results similar to those obtained manually. Secondly, we show that
the reported distribution over divergence types is predictive of the performance
patterns of a zero-shot parser (\S~\ref{sec:zero-shot-parsing}).

\section{Related Work}\label{sec:related_work}

Comparing syntactic and semantic structures over parallel corpora is the subject of much previous work. \citet{dorr2010interlingual} compiled a multiply-parallel corpus and annotated it with increasingly refined categories in an attempt to abstract away from syntactic detail but did not report any systematic measurement of the distribution of divergences. \citet{sindlerova2013verb}, \citet{xue2014not}, \citet{sulem2015conceptual}, and \citet{damonte2018cross} studied divergences over semantic graphs and argument-structure phenomena, while a related line of work examined divergences in discourse phenomena \citep{sostaric-etal-2018-discourse}.
Other works studied the ability of a given grammar formalism to capture CLMD in a parallel corpus \citep[e.g., ][]{sogaard_wu2019}. However, none of these works defined a general methodology for extracting and classifying CLMD.

The only previous work we are aware of to use UD for identifying CLMD is \citep{wong-etal-2017-quantitative}, which addresses Mandarin-Cantonese divergences by comparing the marginal distribution of syntactic categories on both sides (without alignment).
Relatedly, \citet[DX17;][]{deng2017translation} aligned phrase structure trees over an En-Zh parallel corpus. Notwithstanding the similarity in the general approach, we differ from DX17 in (i)~specifically targeting content words, (ii)~relying on UD, which is standardized cross-linguistically and allows to simplify the alignment process by focusing on the level of words,\footnote{Their alignment process involved bottom-up and a top-down passes, sometimes yielding contradictory results.} and (iii)~addressing multiple language pairs.
It should be noted that the classification of divergences presented in DX17 is rather coarse-grained. Of the seven classes in their study, five (Transitivity, Absence of function words, Category mismatch, Reordering, and Dropped elements) reflect local syntactic differences; one (Lexical encoding) covers many-to-one/one-to-many alignments and non-literal word translations; and the remaining residual type (Structural paraphrase) indiscriminately covers more substantial CLMD. We address this limitation and propose a methodology that automatically derives fine-grained CLMD from aligned annotated corpora and enables straightforward computation of their type statistics.

\section{Fine-grained Classification of CLMD}\label{sec:clmd}

In this section, we present a novel cross-linguistic dataset that provides a high-resolution overview of morphosyntactic differences between pairs of languages and a formal definition of morphosyntactic divergences formulated based on it.

Divergences in the syntax of sentences and their translations can stem from a number of reasons. Setting aside semantic divergences, which are differences in the content expressed by the source and the target \citep{carpuat2017detecting,vyas-etal-2018-identifying}, the remaining varieties of divergences are essentially different ways to express the same content \citep{fisiak1984contrastive,boas2010contrastive}, which we call CLMD.

We define CLMD empirically to be recurrent divergence patterns in the syntactic
structures of sentences and their translations. While content differences may
account for some of the observed syntactic divergences, by aiming for recurring
patterns we expect to filter out most such cases, as they are subject to fewer
grammatical constraints and should thus not yield systematic patterns of
morphosyntactic divergence.

It is harder to distinguish between translation artifacts and CLMD in translated sentences that are due to the genuine differences between grammar and usage. However, translated texts are usually characterized by a higher degree of morphosyntactic transfer and rarely portray the target language as more different from the source language than it needs to be \citep{koppel2011translationese,volansky2015features}. Therefore, we do not expect to find spurious recurrent morphosyntactic-divergence patterns introduced by the process of translation.

\subsection{The Manually Aligned PUD Resource}\label{ssec:aligned_pud}

Universal Dependencies (UD) is a framework for treebank annotation, whose objectives include satisfactory analyses of individual languages, providing a suitable basis for bringing out cross-linguistic parallelism, suitability for rapid consistent annotation and accurate automatic parsing, ease of comprehension by non-linguists, and effective support for downstream tasks.
See the Appendix for a glossary of UD terms.

An important feature of the dependency analysis in UD is that content words are considered the principal components of dependency relations. Within this framework, function words are generally dependents of the content word they relate to most closely. 
The primacy of content words brings out cross-linguistic parallels that would be harder to detect with other annotation frameworks since function words are highly variable across languages. Importantly, dependency paths between content words do not generally contain function words. As a result, by comparing paths across languages, differences in the surface realization are often masked, and argument structure and linkage differences emphasized.

For example, a preposition accompanying a verb may be dropped in translation if the corresponding verb is transitive (cf.\ \textit{went around the world} in En vs.\ \textit{oboshel\textsubscript{went.around} mir\textsubscript{world}} in Ru). As prepositions modify the head noun in UD prepositional phrases, the dependency path between the verb and the head noun is not altered.

The Parallel Universal Dependencies (PUD) corpus consists of 1000 sentences
translated into various languages by professional translators.\footnote{Half of
the sentences in the corpus are taken from news articles and the other half from
Wikipedia. 750 of the sentences were originally in English, 100 in German, 50 in
French, 50 in Italian, and 50 in Spanish. All sentences were translated to other
languages via English.}

In this paper, we study the Russian, French, Chinese, Japanese, and Korean versions of the PUD corpus, which were each aligned with the corresponding English corpus.\footnote{Occasionally highly divergent translations prohibited constructing an alignment. 999 sentences were aligned for En-Fr, En-Zh, and En-Jp, 995 for En-Ru, and 884 for En-Ko.} Each parallel corpus was aligned by a human annotator, proficient in the language of the corpus and in English. The UD tokenization is adopted in all cases. Due to the difficulty in finding annotators proficient in pairs of these languages, our annotation takes English as the source language. However, it is possible to obtain an approximate alignment between any pair of these languages, pivoting through English.

Only content words are aligned, so as to sidestep the inherently ambiguous nature of aligning function words across divergent constructions. For details on the function/content distinction we apply to words, see the Appendix.
We restrict the alignment to include connected components of the following types: (1)~one-to-one alignments, i.e., where a single content word is aligned with another single content word; (2)~many-to-one alignments, where multiple source words are aligned with a single target word; (3)~one-to-many alignments, where a single source word is aligned with multiple target words.
Where a source multi-word expression is translated with a target multi-word expression, we align their headwords, to indicate that their subtrees are in correspondence with one another (e.g., English \textit{with this} and French \textit{par cons\'{e}quent}).
Most of the content words in the corpora were aligned in a one-to-one alignment, which accounts for around 90\% of aligned En tokens across the corpora.

\subsection{Defining Divergences using UD}\label{ssec:defining_divergences}

We present a framework for defining and investigating translation divergences across a variety of language pairs using UD. The framework operates on a sentence-aligned parallel corpus, where both sides are annotated with UD and content words in corresponding sentences are aligned.

Let $T_s=(V_s,E_s)$, $T_t=(V_t,E_t)$ be a pair of UD trees over corresponding sentences, and let $CW_s \subset V_s$ and $CW_t \subset V_t$ be the sets of content words in $T_s$ and $T_t$ respectively.
Let $A \subset CW_s \times CW_t$ be a token-level alignment, consisting of one-to-one, many-to-one, and one-to-many alignments.
There are two ways to restrict the definition of correspondences between nodes and edges in $T_s$ and $T_t$: (1)~by considering only one-to-one edges or (2)~by defining a one-to-one correspondence $A' \subset A$ by traversing all many-to-one alignments $C = \{(v_1,u),(v_2,u),...,(v_k,u)\} \subset A$, and selecting for $A'$ only $(v_i,u)$, where $v_i$ is the highest node in $T_s$ among the nodes in $C$. The same is done for one-to-many alignments.\footnote{Cases of non-unique highest nodes are generally rare in PUD and are thus excluded to simplify the analysis. The only frequent case is the Fr discontiguous negation marker \textit{ne... pas}, generally corresponding to \textit{not}.} The first approach is preferable for analyzing syntactic-path correspondences and was followed in this presentation. The second approach is more suitable for analysis of POS mappings, where headwords are more prominent.

We then define \textit{Corresponding Syntactic Relations} (CSR) as a pair $(R_s,R_t)$ such that $R_s$ and $R_t$ are dependency paths in $T_s$ and $T_t$, and such that the origin and endpoint of $R_s$ are in $CW_s$ and the origin and endpoint of $R_t$ are their aligned tokens in $CW_t$ according to $A'$. If the origin or the endpoint of $R_s$ do not have a corresponding node in $T_t$, $R_s$ does not have a corresponding relation in $T_t$. The types of $R_s$ and $R_t$ are the sequence of labels on the edges of the paths, optionally along with their directionality in the tree (linear order is not taken into account).
Without loss of generality, we assume that $R_s$ begins at the leftmost word of the pair in the En sentence, and $R_t$ by definition begins at the target word corresponding to the leftmost source word.
For brevity, we only present results where directionality is not taken into account. Relations are thus written as sequences of UD edge labels separated by the `+' sign.

Token pairs that do not share a POS tag and CSR not of the same type are said to form a divergence. One-to-many and many-to-one alignments are another form of divergence.

\section{Empirical Study of Divergences}\label{sec:corpus_study}

We apply the proposed methodology to the aligned PUD. We compare syntactic relations, analyzing correspondences between POS tags as well as correspondences between single-edge relations in English and target-side dependency paths.

\subsection{Parts of Speech}

We begin by examining the mappings of the POS of corresponding tokens (see the Appendix for the full percentage and count matrices). We find that En POS tags of content words are mostly stable in translations to Fr and Ru (sums of the values on the main diagonals account for 78 and 77\% of the total number of word pairs respectively). Notable exceptions are the negative particle \textit{not}, which is in a one-to-many alignment in French with \textit{ne} and \textit{pas}, certain types of auxiliaries analyzed as verbs in both Ru and Fr, and proper nouns, which often get mapped to Fr nouns (cf.\ \S~\ref{sec:discussion} and discussion in [\citeauthor{samardvzic2010scope}, \citeyear{samardvzic2010scope}]).

The En-Zh matrix presents more divergences with only 65\% of the alignment edges connecting tokens with the same POS. 11\% of nouns were translated as verbs (the reverse mapping is found, albeit to a lesser extent, in all three corpora). Most of such cases involve names of actions and agents (\textit{borrowing}, \textit{ruler}, etc.). En negative particles are split between Zh adverbs, verbs, and auxiliaries; adjectives are quite often mapped to nouns, which form parts of compounds (e.g., \textit{social media} $\rightarrow$ \textit{sh\`{e}ji\={a}o m\'{e}it\v{i}}, lit.\ `social-interaction media'). Adpositions involving spatial relations (the only type of adpositions we consider as content words) are predominantly mapped to adverbs.

The En-Ko matrix is even more divergent: only 62\% of the alignment edges connect matching POS. The most striking property of the En-Ko POS matrix is that NOUN serves as a ``sink'' for other POS: 27\% of En adverbs, 56\% of En adjectives, and 60.5\% of En verbs correspond to Ko nouns. For example, En \textit{trying (to do something)} corresponds to Ko \textit{misu} `attempt'. As we will show in the next section, this is due to drastic syntactic divergences in En-Ko.

The En-Jp matrix is similar: 62.4\% of the edges connect matching POS. Verbs are mostly translated as verbs (58.1\%), which shows more affinity between En and Jp basic clause structure. However, adjectives still mostly turn into nouns (53.7\%), and adverbs are quite likely to get translated by a noun (16.4\% vs.\ 25.8\% for adverb\(\rightarrow\)adverb).

Both Ko and especially Jp tend to leave En pronouns unaligned (15\% and 59\% respectively), upholding their reputations as ``radically pro-drop'' languages \citep{neeleman2007radical}. Interestingly, Zh, another classical example of this phenomenon, loses only 9\% of the pronouns. Ru, a mildly pro-drop language, loses 4\% of the pronouns, while the non-pro-drop Fr loses only 2\%. This demonstrates the fine granularity of distinctions an empirical approach to CLMD can yield.

\subsection{Divergences in Syntactic Relations}\label{par:divergences-relations}

Table \ref{table:ud-divergences} presents the matrices of target-side syntactic relations that correspond to single-edge source-side relations in the five parallel corpora.

Several observations can be made.
First, the En-Fr and En-Ru matrices are similar and are dominated by the elements on the main diagonal (60\% of the total number of edges in En-Fr and 55\% in En-Ru).
An exception are \ttt{compound}s (which in En are mostly noun compounds), as Ru does not have a truly productive nominal compounding process and Fr compounds are annotated as other relations in UD \citep{kahane2017multi}. The other three matrices are less dominated by the entries on the main diagonal (46\% of the alignments in En-Zh, 32\% in En-Ko, 25.8\% in En-Jp) and show higher entropy in most rows, especially in \ttt{nmod}, \ttt{amod}, \ttt{obl}, and \ttt{xcomp}, \ttt{compound} again being a notable exception (entropy matrices for all relations can be found in the Appendix).

Adverbial clauses (\ttt{advcl}) have relatively low values on main diagonals and a high percentage of single edges corresponding to multi-edge paths. This reflects the wide semantic range of \ttt{advcl}: in addition to modifying the matrix predicate (\textit{died \textbf{by drowning}}), they can also denote sequential and parallel events (\textit{published a paper \textbf{sparking a debate}}). The latter two cases naturally give rise to \ttt{conj} and complement clauses (cf.\ \textit{published a paper \textbf{and sparked a debate}} / \textit{published a paper \textbf{to spark a debate}}), the most common other path in En-Ru and En-Fr respectively. As we show in \S~\ref{ssec:nsubj-split}, there is also a converse phenomenon: sequences of events represented using coordinated clauses, \ttt{ccomp}, or \ttt{xcomp} in En are translated with \ttt{advcl} in East Asian languages.

Of particular interest are the differences between En-Ko and En-Jp confusion matrices. Japanese and Korean are largely similar from the point of view of language typology (SOV word order, topic prominence, agglutinative morphology), but there are also important differences on the level of usage. Thus, the adjective class in Korean is less productive, and translations often resort to relative clauses for the purposes of nominal modification. Another difference is the fact that Japanese has few compounds as those are usually translated as \ttt{nmod} with a genitive particle, while Korean translates nearly all En compounds as compounds. See the discussion of further differences in the next section.

\section{Qualitative Analysis of Divergences}\label{sec:refining-ud}

In this section, we analyze prominent cases of divergences revealed by applying our method, attempting to demonstrate how fine-grained CLMD may be detected from the confusion matrices and shedding light on what challenges are involved in bridging these divergences (e.g., for the purposes of MT or cross-lingual transfer). Some of the divergences arise due to real differences between grammars; others are largely due to inconsistent application of the UD methodology.

\subsection{Nominal and Adjectival Modifiers}

When inspecting the translation correspondents of adjectives, we find that while in En-Fr and En-Ru the adjective classes are mostly overlapping, this is not the case for Zh, Ko and Jp. Instead, translation into these languages shifts probability mass from adjectives to nouns: nouns are hardly ever translated to adjectives, but adjectives are more likely to be translated to nouns than remain adjectives. This trend is related to a preference to translate adjectives into possessives (e.g., \textit{Korean company} $\rightarrow$ Jp: {\it Kankoku no kaisha} lit.\ `a company of Korea') or compounds (e.g., \textit{European market} $\rightarrow$ Jp: {\it \={O}sh\={u} ichiba} lit.\ `Europe market').

\subsection{Nominal Subjects}\label{ssec:nsubj-split}

The confusion matrix shows that En \ttt{nsubj} demonstrates very different multi-edge mappings into European languages (Fr and Ru) as opposed to East Asian ones (Zh, Jp, and Ko). The ``most common other path'' for both Russian and French is \ttt{xcomp+nsubj}, which is easy to explain: PUD corpora of these languages ``demote'' fewer auxiliary predicates than English (criteria for demotion are formulated in terms of superficial syntax and differ between languages) and more often place the dependent predicates as the root. Therefore, in constructions like \textit{\textbf{he} could \textbf{do} something} the direct edge between the subject and the verb of the dependent clause is replaced with two edges going through the modal predicate.\footnote{Cf.\ also 23 En \ttt{nsubj} edges mapped to Ru \ttt{nsubj+obl}. Inspection of these sentences reveals that the CLMD can be ascribed to metaphorical usage (e.g., the sense of \textit{read} employed in \textit{the post reads} has no direct correspondent in Ru). Some such cases can be disambiguated using existing annotation schemes.}

In Zh, Ko and Jp, however, there is another issue: sequential events described using coordinated conjuncts and \ttt{xcomp} in En are analyzed as being described with temporal or causal subordinate clauses (\textit{Kipling met and fell in love with Florence Garrard} \(\rightarrow\) Ko: \textit{Kipeulring manna\textsubscript{meet.subordinate} sarange\textsubscript{in.love} ppajyeosseumyeo\textsubscript{fell}}, lit. `having met, fell in love'). This makes the direct \ttt{nsubj} edge in En correspond to an Ko \ttt{nsubj} edge within a subordinated clause, and thus a \ttt{nsubj+advcl} path. Given that not all coordinated verbs are translated using a subordinate clause in Ko and Zh, bridging these divergences is likely to require more than a simple tree-transformation rule but possibly refinement of UD's categories to more abstract linkage types.

\subsection{Modal Auxiliaries in Korean}

UD treats En modal verbs, such as \textit{can} or \textit{may}, as \texttt{aux}, which are dependent on the lexical verb (e.g., could  $\xleftarrow[]{\texttt{aux}}$ do).
Corresponding verbs in other languages are often treated as simple verbs (for example, all Ru modal verbs are simple verbs in UD).

Even more drastically, Ko routinely expresses this semantics by using an existential construction with the literal meaning `(for) X there was a possibility of doing Y' (instead of \textit{X could do Y}), which converts the En \ttt{aux} into \ttt{nsubj}+\ttt{acl}.
In this case, a tree transformation seems to be sufficient to bridge this divergence.

\subsection{\ttt{nmod}\(\rightarrow\)\ttt{acl+X} in Korean}

Ko also differs from other languages in the extent that it uses relative clauses for nominal modification.  Table~\ref{table:ud-divergences} shows that \ttt{nmod} has a high percentage for ``other'' mappings (48\%). Investigation of this long tail shows that to a large extent it consists of \ttt{acl}-based constructions: \ttt{acl+advmod}, \ttt{acl+nsubj}, \ttt{acl+obj}. Added to \ttt{acl}, the cumulative share of \ttt{acl}-based constructions is on par with \ttt{compound}, the main correspondent of this relation for non-possessive \ttt{nmod} (possessive \ttt{nmod} are the only ones that map to \ttt{nmod} in Ko). This discrepancy is due to the fact that Ko nearly obligatorily adds contextually-predictable predicates to oblique relations such as \textit{actions [taken] in Crimea} or \textit{people [being] without children}. The Korean PUD does not demote these verbs to functional-word status (such an approach is advocated for in [\citeauthor{gerdes2016dependency}, \citeyear{gerdes2016dependency}]) but turns them into clause-heading verbs, thus yielding an \ttt{acl+X} divergence.\footnote{The list of examples we can discuss goes on. For example, while investigating the cross-linguistic patterning of English \ttt{advcl}, we noticed that it often gets mapped to \ttt{ccomp} in French and \ttt{acl} in Russian. Both divergent annotations seem to be erroneous as the sentences they appear in are covered by the definition of \ttt{advcl} provided in the UD manual. However, the French case is interesting in that the source \ttt{advcl} in question can be characterized semantically: instead of denoting a secondary action, they reflect a sequence of events or parallel scenes (e.g., \textit{Columbus \textbf{sailed} across the Atlantic... \textbf{sparking} a period of European exploration of the Americas}). Another problem is presented by multi-word expressions analyzed as proper nouns where all tokens have the same POS tag. The UD manual advises to retain the original parts of speech in proper nouns consisting of phrases (e.g., \textit{Cat on a Hot Tin Roof}) but allows to treat \enquote{words that are etymologically adjectives} as PROPN in names such as \textit{the Yellow Pages}. When such names are translated, PROPN get reanalyzed as ADJ, NOUN, etc., producing spurious CLMD.}

\begin{table}[ht]
\centering
\scalebox{.7}{
\begin{tabular}{c@{}c@{}|ccccc}
Language &	 & ru &	fr &	zh &	ko &	jp \\
\hline
\multirow{2}{*}{Thematic}	& Full	                      & 25 &	4	&   8	&   5	 & 5 \\
	& \ttt{nsubj} to \ttt{obj}/\ttt{obl}                  &	78 &	57  &	43 &	25 &   53 \\
Promotional &	                                          &	0 &	    0 &     0 & 	0  &   0 \\
Demotional	&	                                          & 10 & 	2 &	    4 &	    19 &   1 \\
Structural	 &	                                          & 83 &	67 &	17 &	0 &	   35 \\
Conflational &                                            &	10 &	5 &	    5 &	    6 &	   2 \\
\multirow{2}{*}{Categorical} &	\ttt{nsubj}+\ttt{obj}     &	8 &	    12 &	23 &	11 	&  4 \\
 &	\,\ttt{nsubj}+\ttt{(i)obj/obl}\,	                  & 51 &	34 &	25 &    11	 & 13 \\
\#Sentences &	                                          &	995 & 	999 &	999 & 	884 &	999 \\
\end{tabular}}
\caption{Prevalence in absolute counts of translation divergences as defined in \newcite{dorr1994machine}. Row headings are explained in  \S~\ref{sec:dorr_td}. \label{tab:dorr_td}
}
\end{table}

\section{Revisiting Dorr's Divergences}\label{sec:dorr_td}

We turn to show that the types of the seminal classification of divergences from \newcite{dorr1994machine} can be, with a single exception, recast in terms of UD divergences. We quote the original formulations of the divergences illustrated through English-Spanish or English-German examples.

\paragraph{Thematic divergence:} E: \textit{I like Mary} $\Leftrightarrow$ S:~\textit{Mar\'{i}a me gusta a m\'{i}} `Mary pleases me.' In UD, this corresponds to the situation when the original \ttt{obj} or \ttt{obl} becomes the \ttt{nsubj} and vice versa. The divergence will correspond to a CSR of type $(nsubj,obj)$ or $(nsubj,obl)$. A ``full'' thematic divergence will also involve the inverse divergence $(obj,subj)$ or $(obl,subj)$.

\paragraph{Promotional divergence:} E: \textit{John usually goes home} $\Leftrightarrow$ S:~\textit{Juan suele ir a casa} `John tends to go home.' This corresponds to the situation where the original root predicate becomes an \ttt{xcomp}, and the original \ttt{advmod} takes its place as the root. Corresponding CSR type: $(advmod,xcomp)$.

\paragraph{Demotional divergence:} E: \textit{I like eating} $\Leftrightarrow$ G:~\textit{Ich esse gern} `I eat likingly.' The original \ttt{xcomp} becomes the root predicate, and the original root predicate is demoted to the position of an \ttt{advmod}. The relevant CSR type is $(xcomp,advmod)$.

\paragraph{Structural divergence:} E: \textit{John entered the house} $\Leftrightarrow$ S:~\textit{Juan entr\'{o} en la casa} `John entered in the house.' The original \ttt{obj} becomes an \ttt{obl}. CSR type: $(obj,obl)$.

\paragraph{Conflational divergence:} E:~\textit{I stabbed John} $\Leftrightarrow$ S:~\textit{Yo le di pu\~{n}aladas a Juan} `I gave knife-wounds to John.' The original root predicate is in a one-to-many alignment with a combination of a root predicate and its \ttt{obj}.

\paragraph{Categorial divergence:} E: \textit{I am hungry} $\Leftrightarrow$ G:~\textit{Ich habe Hunger} `I have hunger.' The original root predicate becomes an \ttt{obj}. CSR type: $(nsubj,nsubj+obj)$.

\paragraph{Lexical divergence:} E: \textit{John broke into the room} $\Leftrightarrow$ S:~\textit{Juan forz\'{o} la entrada al cuarto} `John forced (the) entry to the room.'
Divergences of this type arise whenever aligned words have at best partially overlapping semantic content and never appear on their own but always with other divergences. The information necessary to ascertain the degree of word-meaning overlap is not embedded into UD or any other cross-lingual annotation scheme; therefore we were unable to provide a formal interpretation of this type of divergence.

\paragraph{Frequencies of Dorr's Divergences in PUD}
are presented in Table~\ref{tab:dorr_td} (except for Lexical divergences, which are hard to formalize). It is evident that these types only account for a small portion of the encountered divergences, the point already made for En-Zh in DX17. It seems then that ``hand-crafted'' translation divergences, however insightful they may be, receive attention disproportionate to their empirical frequency.

\section{Perspectives for Automation}\label{sec:automation}

One of the strengths of our approach is that it only relies on UD parses and alignments, for which automatic tools exist in many languages. To demonstrate the feasibility of an automated protocol, we conducted an analysis of the WMT14 En-Zh News Commentary corpus.\footnote{\url{http://www.statmt.org/wmt14/}} We used TsinghuaAligner \citep{liusun2015aligner} and pretrained English and Chinese UD parsers from the StanfordNLP toolkit \citep{qi2018universal}. To verify that the aligner we are using is adequate for the task, we aligned the En-Zh PUD corpus pair and checked the resulting precision and recall of the edges corresponding to content words.\footnote{These were defined here as edges with the following labels: \texttt{root}, \texttt{nsubj}, \texttt{amod}, \texttt{nmod}, \texttt{advmod}, \texttt{nummod}, \texttt{acl}, \texttt{advcl}, \texttt{xcomp}, \texttt{compound}, \texttt{flat}, \texttt{obj}, \texttt{obl}.} The results (P=\(0.86\), R=\(0.32\)) indicate that the automated approach is able to recover around a third of the information obtained through manual alignment with reasonable precision. Importantly, we find recall to be nearly uniform for all source edge types, which suggests that the low recall can be mitigated by using a larger corpus without biasing the results.

The POS and edge-type confusion matrices built from this experiment are very similar to the ones reported in this paper (save for \texttt{compound}, which is not produced by the Stanford Zh parser), and are not reproduced here (they can be found in the Supplementary Materials\footnote{\url{https://www.cse.huji.ac.il/~oabend/data/translation_divergences_supp.zip}}).

\section{Applicability for Zero-Shot Parsing}\label{sec:zero-shot-parsing}

We come to demonstrate the applicability of our method for analyzing the performance of a downstream cross-lingual transfer task.
We consider zero-shot cross-lingual parsing  \citep{ammar-etal-2016-many,schuster-etal-2019-cross} as a test case and investigate to what extent the performance of a zero-shot parser on a given dependency label can be predicted from its stability in translation.
As test sets we use the test sets of GSD UD corpora for the five languages (Ru, Fr, Zh, Ko, and Jp), as well as the corresponding PUD corpora.
We train a parser following the setup of \citet{mulcaire-etal-2019-low} and use a pretrained multilingual BERT \citep{devlin2018bert}, feeding its output embeddings into a biaffine-attention neural UD parser \citep{dozat2017biaffine} trained on the English EWT corpus. We evaluate the parser's ability to predict relation types by computing F-scores for each dependency label (save for labels corresponding to function words that were generally not aligned). The Appendix gives full implementation details.

We start by computing Spearman correlations between F-scores and the \textsc{preservation} indices, defined as the proportion of identity mappings in the confusion matrices for each corpus (e.g., {\sc preservation} for \ttt{acl} in Ru is 0.48, while in Jp it is 0.37). The correlations are very strong for some languages and noticeable for others ($\rho = 0.62, 0.75, 0.31, 0.42, 0.77$ for Ru, Fr, Zh, Ko, and Jp respectively on GSD test sets, and $\rho = 0.7, 0.82, 0.72, 0.84, 0.68$ on PUD).

We hypothesize that the preservation of a relation in translation is related to the ability of a zero-shot parser to predict it. In order to control for obvious covariates, we introduce two control variables: (1)~\textsc{source-side hardness} (test-set F-scores attained by the parser on English dependency relations) and (2)~\textsc{target-side hardness} (F-scores attained by a parser trained on the target-language UD GSD corpus on the target-language test set). We use a mixed-effects model with \textsc{preservation}, \textsc{source-side hardness}, and \textsc{target-side hardness} as fixed effects, random intercepts for language, and F-scores for dependency relations as the dependent variable.
We then used likelihood-ratio test to compute \textit{p}-values for the difference in predictive power between the model without \textsc{preservation} and one with it. The \textit{p}-value (using Holm correction) is highly significant ($<$ 0.001) for the PUD corpora, and for GSD it is significant with $p=0.02$.

These results suggest that morphosyntactic differences between languages, as uncovered by our method, play a role in the transferability of parsers across languages. This also underscores the potential utility of bridging CLMD for improving syntactic transfer across languages.

\section{Discussion}\label{sec:discussion}

The presented methodology gives easy access to different levels of analysis. On one hand, by focusing on content words, the approach abstracts away from much local-syntactic detail (such as reordering or adding/removing function words).
At the same time, the methodology and datasets provide means to investigate essentially any kind of well-defined CLMD.
Indeed, since function words in UD tend to be dependents of content words, we may analyze the former by considering the distribution of function word types that each type of content word has.
Moreover, sub-typing dependency paths based on their linear direction can allow investigating word-order differences.\footnote{Surface-Syntactic Universal Dependencies may be a better fit for such type of analysis, see \cite{gerdes2018sud}.}

 Other than informing the development of cross-lingual transfer learning, our analysis directly supports the validation of UD annotation.
 For example, we reveal inconsistencies in the treatment of multi-word expressions across languages. Thus, the translation of many NPs with adjectival modifiers, such as \textit{Celtic sea} or \textit{episcopal church}, are analyzed as \ttt{compound}s. Languages such as Ru, lacking a truly productive nominal-compound relation, carve this class up based mostly on the POS of the dependent element (e.g., \textit{episcopal} corresponds to a Ru \ttt{amod}), its semantic class (e.g., compounds with cardinal directions are Ru \ttt{amod}s), and whether the dependent element itself has dependents (these mostly correspond to Ru \ttt{nmod}s).  Our method can be used to detect and bridge such inconsistencies.

In conclusion we note that our analysis suggests that considerable entropy in the mapping between the syntactic relations of the source and target sides can be reduced by removing inconsistencies in the application of UD, and perhaps more importantly by refining UD with semantic distinctions that will normalize corresponding constructions across languages to have a similar annotation. This will simultaneously advance UD's stated goal of ``bringing out cross-linguistic parallelism across languages'' and, as our results on zero-shot parsing suggest, make it more useful for cross-linguistic transfer.

\section*{Acknowledgments}

We thank Nathan Schneider for helpful comments and anonymous reviewers for useful feedback. This work was supported by the Israel Science Foundation (grant no. 929/17).

\begin{table*}[t]
\centering
\scalebox{0.53}{
\begin{tabular}{lllllllllllllllllllllll}
 &  &
\rotatebox{90}{\textbf{acl}} &
\rotatebox{90}{\textbf{advcl}} &
\rotatebox{90}{\textbf{advmod}} &
\rotatebox{90}{\textbf{amod}} &
\rotatebox{90}{\textbf{appos}} &
\rotatebox{90}{\textbf{ccomp}} &
\rotatebox{90}{\textbf{compound}} &
\rotatebox{90}{\textbf{conj}} &
\rotatebox{90}{\textbf{fixed}} &
\rotatebox{90}{\textbf{flat}} &
\rotatebox{90}{\textbf{nmod}} &
\rotatebox{90}{\textbf{nsubj}} &
\rotatebox{90}{\textbf{nummod}} &
\rotatebox{90}{\textbf{obj}} &
\rotatebox{90}{\textbf{obl}} &
\rotatebox{90}{\textbf{parataxis}} &
\rotatebox{90}{\textbf{xcomp}} &
\rotatebox{90}{\textbf{Collapsed}} &
\rotatebox{90}{\textbf{Other}} &
\rotatebox{90}{\textbf{MCOP (\%)}} &
\rotatebox{90}{\textbf{MCOP}} \\
\multirow{17}{*}{\rotatebox{90}{\Large English-Russian}} & \textbf{acl} & 48 & 1 & & 2 & & 1 & & & & & 9 & 2 & & 2 & 1 & 1 & 4 & 1 & 29 & 5 & nmod+acl \\
& \textbf{advcl} & 5 & 32 & 1 & & & 4 & & 5 & & & 1 & & & & 10 & 2 & 1 & & 37 & 2 & advcl+xcomp \\
& \textbf{advmod} & & & 63 & 2 & & & & & & & 1 & & & & 3 & & & 5 & 24 & 2 & advmod+nummod \\
& \textbf{amod} & 3 & & 1 & 77 & & & & & & & 5 & & 1 & & & & & 5 & 7 & 2 & det \\
& \textbf{appos} & & & & & 36 & & & 1 & & 13 & 18 & 1 & 1 & & & 1 & & 1 & 29 & 6 & nmod+flat \\
& \textbf{ccomp} & 1 & 2 & 1 & & & 49 & & & & 1 & & & & 1 & & 6 & 6 & & 33 & 11 & ccomp+xcomp \\
& \textbf{compound} & & & & 33 & 1 & & 1 & & & 6 & 23 & & 1 & & 1 & & & 20 & 13 & 3 & nmod+nmod \\
& \textbf{conj} & & & & & & & & 76 & & & & & & & & & & 1 & 21 & 2 & conj+conj \\
& \textbf{fixed} & & & & & & & & & & 15 & 8 & & & & & & & 69 & 8 & 8 & nmod+flat \\
& \textbf{flat} & & & & 3 & 1 & & 1 & & & 75 & 6 & & & & & & & 4 & 9 & 3 & flat+flat \\
& \textbf{nmod} & & & 1 & 4 & 1 & & & & & 1 & 59 & & 1 & 1 & 3 & & & & 29 & 6 & det \\
& \textbf{nsubj} & & & & & & & & & & & 1 & 71 & & 2 & 2 & & & & 22 & 4 & nsubj+xcomp \\
& \textbf{nummod} & & & 1 & 5 & & & & & & 1 & 2 & & 82 & & 1 & & & 7 & 2 & 1 & compound+nummod \\
& \textbf{obj} & & & 1 & 1 & & & & & & & 8 & 4 & & 57 & 9 & & & 4 & 15 & 4 & iobj \\
& \textbf{obl} & & 1 & 5 & & & & & & & & 4 & 2 & & 3 & 50 & & & 1 & 34 & 7 & iobj \\
& \textbf{parataxis} & & 1 & & & & 1 & & 7 & & & & & & & & 59 & & & 31 & 3 & acl+obl \\
& \textbf{xcomp} & & 3 & 5 & & & 4 & & 1 & & & 1 & 2 & & 2 & 8 & & 51 & 2 & 22 & 8 & iobj \\
\hline
& & &  &  &  &  &  &  &  &  &  &  &  &  &  &  &  &  &  &  &  &  \\
\multirow{17}{*}{\rotatebox{90}{\Large English-French}} & \textbf{acl} & 34 & 1 & & 4 & 1 & 17 & & 1 & & & 4 & & & 1 & 1 & & 8 & 1 & 26 & 3 & acl+xcomp \\
& \textbf{advcl} & 1 & 44 & 1 & & & 9 & & 2 & & & & & & & 7 & 2 & 2 & & 32 & 4 & xcomp+advcl \\
& \textbf{advmod} & & & 63 & 1 & & & & & & & 2 & & & 1 & 3 & & 1 & 6 & 21 & 4 & advmod+xcomp \\
& \textbf{amod} & & & 1 & 75 & & 1 & & & & & 9 & & & & & & & 5 & 7 & 1 & det \\
& \textbf{appos} & & & & 2 & 69 & 1 & & 1 & & & 6 & & & & 1 & & & & 21 & 3 & appos+nmod \\
& \textbf{ccomp} & & 1 & & & & 52 & & & & & & & & 2 & & 5 & 11 & & 29 & 10 & ccomp+xcomp \\
& \textbf{compound} & & & & 20 & 4 & & 7 & 1 & & 3 & 42 & & 2 & 1 & 1 & & & 14 & 6 & 1 & nmod+nmod \\
& \textbf{conj} & & & & & & & & 77 & & & 1 & & & & & 1 & & 1 & 18 & 2 & nmod+conj \\
& \textbf{fixed} & & & & 17 & 17 & & & & & 17 & & & & & & & & 17 & 33 & 17 & appos+nummod \\
& \textbf{flat} & & & & 2 & 15 & & 2 & & & 65 & 2 & & 4 & & & & & 4 & 7 & 3 & appos+flat \\
& \textbf{nmod} & & & 1 & 2 & & & & & & & 66 & & 1 & 1 & 3 & & & 1 & 24 & 3 & nmod+nmod \\
& \textbf{nsubj} & & & & & & & & & & & 1 & 79 & & 1 & 1 & & & & 18 & 5 & nsubj+xcomp \\
& \textbf{nummod} & & & & 1 & 1 & & & & & & 17 & & 71 & & & & & 3 & 5 & 3 & det \\
& \textbf{obj} & & & & & & & & & & & 5 & 1 & & 72 & 8 & & 1 & 3 & 9 & 2 & obj+nmod \\
& \textbf{obl} & & & 2 & & & & & & & & 2 & 1 & & 6 & 66 & & 1 & 1 & 20 & 3 & obl+nmod \\
& \textbf{parataxis} & 1 & & 1 & & & 5 & & 12 & & & & & & & & 49 & & & 31 & 3 & obl+parataxis \\
& \textbf{xcomp} & & 3 & 3 & & 1 & 2 & & & & & 1 & 1 & & 3 & 4 & & 71 & 4 & 9 & 2 & obj+amod \\
\hline
& & &  &  &  &  &  &  &  &  &  &  &  &  &  &  &  &  &  &  &  &  \\
\multirow{17}{*}{\rotatebox{90}{\Large English-Chinese}} & \textbf{acl} & 32 & 1 & & 2 & & & 2 & & & & 1 & 8 & & 2 & 1 & & 1 & 1 & 48 & 3 & obj+advcl \\
& \textbf{advcl} & & 21 & & & & 2 & & & & & & & & & 1 & & 17 & & 57 & 7 & dep \\
& \textbf{advmod} & 1 & 1 & 38 & 1 & & & 1 & & & & & 1 & & 3 & 4 & & 2 & 9 & 38 & 3 & advmod+obj \\
& \textbf{amod} & 4 & & 2 & 17 & & & 31 & & & & 5 & 3 & 3 & & 1 & & & 14 & 18 & 3 & compound+compound \\
& \textbf{appos} & 1 & & & & 50 & & 8 & 2 & & 2 & 5 & 2 & & & & & 1 & 3 & 27 & 2 & compound+compound \\
& \textbf{ccomp} & & 1 & 2 & & & 37 & & & & & & & & & & & 5 & & 55 & 4 & ccomp+advcl \\
& \textbf{compound} & & & & 2 & & & 42 & 1 & & 2 & 3 & 1 & & & & & & 33 & 14 & 8 & compound+compound \\
& \textbf{conj} & & 10 & & & & & 1 & 45 & & & & & & & & & & 2 & 40 & 5 & dep \\
& \textbf{fixed} & & & & & 2 & & & & & 2 & 2 & & & & & & & 94 & 2 & 2 & appos+flat \\
& \textbf{flat} & & & & & 9 & & 9 & 1 & & 47 & & & & & & & & 9 & 25 & 18 & appos+flat \\
& \textbf{nmod} & 2 & & & 1 & 1 & & 18 & & & & 24 & 3 & 1 & 4 & 3 & & & 3 & 39 & 3 & compound+compound \\
& \textbf{nsubj} & & & & & & & & & & & & 60 & & 3 & 1 & & & & 35 & 8 & nsubj+advcl \\
& \textbf{nummod} & & & & 1 & & & & & & 1 & & & 43 & & & & & 23 & 33 & 24 & nummod+clf \\
& \textbf{obj} & 1 & 1 & & & & 1 & 1 & & & & & 1 & & 59 & 4 & & 1 & 4 & 27 & 7 & ccomp+nsubj \\
& \textbf{obl} & & 1 & 2 & & & 2 & 1 & & & & 1 & 4 & & 16 & 19 & & 2 & 2 & 50 & 6 & obj+advcl \\
& \textbf{parataxis} & & 2 & 1 & & & 7 & & & & & & & & & & & 1 & & 88 & 22 & dep \\
& \textbf{xcomp} & 1 & 3 & 7 & & 1 & 15 & & & & & & 1 & & 6 & 1 & & 26 & 3 & 38 & 13 & aux \\
\hline
 & &  &  &  &  &  &  &  &  &  &  &  &  &  &  &  &  &  &  &  &  &  \\
\multirow{17}{*}{\rotatebox{90}{\Large English-Korean}} & \textbf{acl} & 37 & 2 & 2 & & & & 2 & & & & 1 & 1 & & 2 & & & & & 54 & 4 & acl+nsubj+acl \\
& \textbf{advcl} & 2 & 18 & 7 & & & & & & & 1 & & 2 & & & & & & & 71 & 6 & advmod+acl \\
& \textbf{advmod} & 1 & 2 & 43 & 1 & & & 3 & & & & & 1 & & 1 & & & & 4 & 43 & 8 & aux \\
& \textbf{amod} & 8 & & 2 & 12 & & & 38 & & & 1 & 8 & 2 & 2 & & & & & 10 & 19 & 5 & det \\
& \textbf{appos} & 27 & 1 & & & 8 & & 17 & & & 3 & 2 & 2 & & & & & & & 40 & 4 & acl+compound \\
& \textbf{ccomp} & & 30 & 3 & & & 2 & & & & 2 & & & & & & & & & 64 & 8 & advcl+advcl \\
& \textbf{compound} & & & 1 & & & & 57 & 1 & & 3 & 2 & 1 & & 1 & & & & 23 & 11 & 2 & compound+compound \\
& \textbf{conj} & 1 & 12 & 1 & & & & 1 & 53 & & 1 & 1 & & & & & & & 1 & 31 & 2 & conj+compound \\
& \textbf{fixed} & & & & & & & 29 & & & & & & & & & & & 43 & 29 & 14 & compound+compound \\
& \textbf{flat} & 4 & & & & & & 17 & & & 60 & 1 & 1 & & & & & & 5 & 13 & 3 & compound+flat \\
& \textbf{nmod} & 1 & & 3 & & 1 & & 17 & & & & 25 & 3 & 1 & 1 & & & & 1 & 45 & 6 & acl+advmod \\
& \textbf{nsubj} & 1 & & 2 & & & & 1 & & & & & 47 & & 3 & & & & & 45 & 7 & nsubj+advcl \\
& \textbf{nummod} & & & 1 & & & & 2 & & & & & & 68 & & & & & 1 & 27 & 10 & nummod+nmod \\
& \textbf{obj} & 1 & 1 & 6 & & & & 3 & & & & & 4 & & 52 & & & & 2 & 30 & 6 & aux+obj \\
& \textbf{obl} & 1 & 1 & 32 & & & & 2 & & & & & 4 & & 5 & & & & 1 & 54 & 4 & advmod+compound \\
& \textbf{parataxis} & 2 & 21 & & & & & & & & & & & & & & & & & 77 & 12 & advcl+advcl \\
& \textbf{xcomp} & 2 & 10 & 14 & & & 7 & 7 & & & & & 1 & & 6 & & & & 2 & 50 & 9 & aux \\
\hline
 & & &  &  &  &  &  &  &  &  &  &  &  &  &  &  &  &  &  &  &  &  \\
\multirow{17}{*}{\rotatebox{90}{\Large English-Japanese}} & \textbf{acl} & 33 & 4 & & & & & 1 & & & & 7 & 1 & & 1 & & & & 4 & 45 & 5 & nmod+acl \\
& \textbf{advcl} & & 15 & & & & & 4 & & & & & & & 4 & & & & 15 & 56 & 7 & obj+acl \\
& \textbf{advmod} & 1 & 2 & 15 & 1 & & & 4 & & & & 1 & 2 & 2 & 1 & 2 & & & 13 & 46 & 9 & aux \\
& \textbf{amod} & 9 & & & 8 & & & 22 & & & & 16 & 1 & 2 & 1 & & & & 27 & 13 & 2 & compound+compound \\
& \textbf{appos} & 3 & 2 & & & 9 & & 18 & & & & 21 & & & & & & & 2 & 34 & 11 & nmod+nmod \\
& \textbf{ccomp} & 7 & 7 & & & & 13 & & & & & & & & & 7 & & & & 60 & 7 & aux \\
& \textbf{compound} & & & & 1 & & & 31 & & & & 15 & & & & & & & 39 & 10 & 3 & compound+compound \\
& \textbf{conj} & 1 & 5 & & 1 & & & & & & & 31 & & & & 1 & & & 2 & 48 & 13 & nmod+nmod \\
& \textbf{fixed} & & & & & & & 2 & & & & 2 & & & & & & & 90 & 4 & 2 & case \\
& \textbf{flat} & & & & & & & 14 & & & & 41 & & & & & & & 18 & 20 & 7 & nmod+compound \\
& \textbf{nmod} & & & & & & & 8 & & & & 40 & & & 1 & 1 & & & 7 & 35 & 7 & nmod+nmod \\
& \textbf{nsubj} & & & & & & & 1 & & & & 2 & 43 & & 1 & 2 & & & 5 & 39 & 5 & nsubj+advcl \\
& \textbf{nummod} & 1 & & & & & & 4 & & & & & & 52 & & 1 & & & 8 & 21 & 13 & nummod+compound \\
& \textbf{obj} & & 1 & & & & & 4 & & & & 6 & 8 & & 44 & 1 & & & 10 & 21 & 5 & iobj \\
& \textbf{obl} & & & 2 & & & & 2 & & & & 4 & 1 & & 3 & 18 & & & 3 & 53 & 14 & iobj \\
& \textbf{parataxis} & & 6 & & & & 6 & & & & & & & & & & & & & 75 & 13 & advcl+obl \\
& \textbf{xcomp} & 7 & 11 & 4 & & & 7 & & & & & & 2 & & 4 & 2 & & & 4 & 43 & 15 & aux
\end{tabular}%
}
\caption{Percentages of corresponding syntactic relations of UD edge types connecting content words in the five corpora. Rows correspond to En and sum to 100\%; columns correspond to target side paths.
``Collapsed'' are cases where the edge is collapsed to a single node.
``MCOP'' stands for most common other target-side path. See the Appendix for raw counts.}
\label{table:ud-divergences}
\end{table*}

\FloatBarrier
\bibliography{Main}
\bibliographystyle{acl_natbib}

\end{document}


\appendix

\noindent \textbf{\Large{Appendices}}

\section{Glossary of UD POS tags and edge types}\label{ssec:ud-glossary}

A glossary of UD POS labels and edge types mentioned in the paper is provided in Table~\ref{tab:categories}.

\begin{table*}[ht]
\centering
\small
\begin{tabular}{|c|p{10cm}|}
\hline
Label & Short Definition\\
\hline
\hline
\multicolumn{2}{|c|}{{\bf POS tags}}\\
\hline
{\it NOUN} & Noun.\\
\hline
{\it PROPN} & Proper noun.\\
\hline
{\it PRON} & Pronoun.\\
\hline
{\it ADJ} & Adjective.\\
\hline
{\it ADV} & Adverb.\\
\hline
{\it VERB} & Verb.\\
\hline
{\it AUX} & Auxiliary.\\
\hline
{\it PART} & Particle.\\
\hline
{\it NUM} & Numeral.\\
\hline
{\it ADP} & Adposition.\\
\hline
\multicolumn{2}{|c|}{{\bf Clause Elements}}\\
\hline
 {\it nsubj} & Nominal subject.\\
\hline
 {\it dobj} & Direct object.\\
\hline
 {\it ccomp} & Clausal complement (finite or infinite), unless its subject is controlled.\\
\hline
 {\it xcomp} & Open clausal complement, i.e., predicative or clausal complement without its own subject.\\
 \hline
 {\it advmod} & Modifying adverb.\\
 \hline
 {\it neg} & Negation modifier (e.g., ``not'', ``no'').\\
\hline
 {\it aux} & Auxiliary of a verbal predicate, including markers of tense, mood, modality, aspect, voice or evidentiality.\\
\hline
 {\it nmod} & Oblique: nominal functioning as an adjunct. ({\it nmods } are also used for nominal modifiers in noun phrases, see below)\\
\hline \hline
\multicolumn{2}{|c|}{{\bf Inter-clause Linkage}}\\
\hline
 {\it conj} & Relation between the conjuncts in a coordination to the first conjunct, which is considered the head.\\
\hline
 {\it cc} & Coordinating conjunction.\\
\hline
 {\it advcl} & Adverbial clause modifier, including temporal clause, consequence, conditional clause, and purpose clause.\\
\hline
 {\it mark} & Marker: the word introducing a clause subordinate to another clause, often a subordinating conjunction. \\
\hline
 {\it parataxis} & Several elements (often clauses or fragments) placed side by side without any explicit coordination, subordination, or argument relation.\\
 \hline \hline

\multicolumn{2}{|c|}{{\bf Nominal Elements}}\\
\hline
 {\it det} & Determiner.\\
\hline
 {\it case} & Case marker, including adpositions.\\
\hline
 {\it nmod} & Nominal modifier of a noun or a noun phrase.\\
\hline
 {\it nummod} & Numeric modifier.\\
\hline

\end{tabular}
\caption{UD POS tags and edge types mentioned in the paper and their definitions.}\label{tab:categories}
\end{table*}

\section{Distinguishing Content and Function Words}\label{app:content-function}

We adopt the following distinction between function and content words. Function words include (1) grammatical-relation markers (prepositions marking direct and indirect objects, possession, and other types of relations inside NPs); (2) tense-aspect-mood markers including inflected auxiliaries; (3) markers of (in)definiteness; (4) coordinating conjunctions; (5) complementizers; (6) classifiers; (7) copulas and existential predicates; (8) dummy subjects and expletives.
Other word types are considered content words, including (1) all other types of predicates, participants, obliques, adverbial and adjectival modifiers; (2) negation markers; (3) discourse markers; (4) quantifiers; (5) spatial/temporal-relations markers.

\section{Confusion matrices for translation equivalents of POS tags and UD edge labels: raw counts}

\subsection{POS tags}\label{sssec:appendix:pos}

Raw-count and percentage confusion matrices for translation equivalents of UD POS tags in three parallel corpora are presetned in Tables~\ref{table:ud-divergences-pos-raw} and~\ref{table:ud-divergences-pos-percent}.

\begin{table*}[t]
\centering
\scalebox{.55}{%
\begin{tabular}{llllllllllllllllll}
	\textbf{En\textbackslash Ru} & \textbf{} & \textbf{ADJ} & \textbf{ADP} & \textbf{ADV} & \textbf{AUX} & \textbf{CCONJ} & \textbf{DET} & \textbf{NOUN} & \textbf{NUM} & \textbf{None} & \textbf{PART} & \textbf{PRON} & \textbf{PROPN} & \textbf{SCONJ} & \textbf{SYM} & \textbf{VERB} & \textbf{X} \\
	\textbf{} & \textbf{ADJ} & 1030 & 1 & 52 & 0 & 0 & 28 & 74 & 2 & 29 & 2 & 0 & 22 & 0 & 0 & 57 & 7 \\
	\textbf{} & \textbf{ADP} & 4 & 296 & 11 & 0 & 0 & 2 & 12 & 0 & 9 & 1 & 0 & 0 & 1 & 0 & 7 & 3 \\
	\textbf{} & \textbf{ADV} & 31 & 10 & 362 & 0 & 25 & 5 & 17 & 2 & 31 & 28 & 1 & 1 & 0 & 0 & 6 & 12 \\
	\textbf{} & \textbf{AUX} & 7 & 0 & 2 & 5 & 0 & 0 & 1 & 0 & 3 & 0 & 0 & 0 & 0 & 0 & 80 & 0 \\
	\textbf{} & \textbf{CCONJ} & 0 & 0 & 0 & 0 & 25 & 0 & 0 & 0 & 2 & 0 & 0 & 0 & 0 & 0 & 0 & 0 \\
	\textbf{} & \textbf{DET} & 31 & 1 & 3 & 0 & 0 & 29 & 3 & 9 & 4 & 8 & 1 & 0 & 0 & 0 & 1 & 0 \\
	\textbf{} & \textbf{INTJ} & 0 & 0 & 0 & 0 & 0 & 0 & 0 & 0 & 0 & 0 & 0 & 0 & 0 & 0 & 0 & 1 \\
	\textbf{} & \textbf{NOUN} & 204 & 3 & 23 & 0 & 0 & 14 & 3013 & 4 & 59 & 0 & 8 & 116 & 0 & 5 & 38 & 12 \\
	\textbf{} & \textbf{NUM} & 20 & 0 & 1 & 0 & 0 & 1 & 23 & 229 & 2 & 0 & 1 & 0 & 0 & 0 & 0 & 4 \\
	\textbf{} & \textbf{None} & 102 & 44 & 90 & 1 & 10 & 38 & 387 & 6 & 0 & 27 & 61 & 32 & 1 & 0 & 177 & 14 \\
	\textbf{} & \textbf{PART} & 0 & 1 & 1 & 0 & 0 & 0 & 0 & 0 & 0 & 45 & 0 & 0 & 0 & 0 & 3 & 0 \\
	\textbf{} & \textbf{PRON} & 5 & 0 & 2 & 0 & 0 & 127 & 13 & 0 & 22 & 0 & 348 & 1 & 0 & 0 & 1 & 0 \\
	\textbf{} & \textbf{PROPN} & 144 & 0 & 0 & 0 & 0 & 3 & 124 & 0 & 9 & 0 & 0 & 1247 & 0 & 0 & 2 & 10 \\
	\textbf{} & \textbf{SCONJ} & 0 & 13 & 3 & 0 & 0 & 0 & 0 & 0 & 0 & 3 & 0 & 0 & 0 & 0 & 0 & 0 \\
	\textbf{} & \textbf{SYM} & 0 & 0 & 0 & 0 & 0 & 0 & 2 & 0 & 0 & 0 & 0 & 0 & 0 & 15 & 0 & 0 \\
	\textbf{} & \textbf{VERB} & 52 & 7 & 9 & 9 & 0 & 0 & 145 & 0 & 66 & 0 & 1 & 2 & 0 & 0 & 1444 & 0 \\
	\textbf{} & \textbf{X} & 2 & 0 & 0 & 0 & 0 & 0 & 2 & 0 & 0 & 0 & 0 & 6 & 0 & 0 & 0 & 0 \\
	\textbf{En\textbackslash Fr} & \textbf{} &  &  &  &  &  &  &  &  &  &  &  &  &  &  &  &  \\
	\textbf{} & \textbf{ADJ} & 1006 & 12 & 51 & 0 & 0 & 2 & 126 & 1 & 44 & 0 & 0 & 12 & 0 & 0 & 45 & 2 \\
	\textbf{} & \textbf{ADP} & 5 & 388 & 21 & 0 & 1 & 4 & 13 & 0 & 9 & 0 & 0 & 0 & 1 & 0 & 7 & 0 \\
	\textbf{} & \textbf{ADV} & 30 & 28 & 416 & 1 & 5 & 2 & 33 & 0 & 43 & 0 & 5 & 0 & 5 & 0 & 7 & 0 \\
	\textbf{} & \textbf{AUX} & 0 & 0 & 2 & 4 & 0 & 0 & 0 & 0 & 2 & 0 & 0 & 0 & 0 & 0 & 74 & 0 \\
	\textbf{} & \textbf{CCONJ} & 0 & 1 & 12 & 0 & 482 & 0 & 0 & 0 & 29 & 0 & 0 & 0 & 0 & 1 & 0 & 0 \\
	\textbf{} & \textbf{DET} & 25 & 2 & 4 & 0 & 0 & 19 & 4 & 5 & 5 & 0 & 2 & 0 & 0 & 0 & 0 & 0 \\
	\textbf{} & \textbf{INTJ} & 0 & 0 & 0 & 0 & 0 & 0 & 0 & 0 & 0 & 0 & 0 & 0 & 0 & 0 & 1 & 0 \\
	\textbf{} & \textbf{NOUN} & 113 & 9 & 9 & 0 & 0 & 1 & 3244 & 10 & 46 & 0 & 10 & 13 & 0 & 5 & 38 & 1 \\
	\textbf{} & \textbf{NUM} & 8 & 0 & 0 & 0 & 0 & 11 & 18 & 371 & 1 & 0 & 1 & 0 & 0 & 0 & 0 & 0 \\
	\textbf{} & \textbf{None} & 55 & 28 & 70 & 3 & 52 & 5 & 162 & 3 & 0 & 0 & 57 & 3 & 0 & 0 & 170 & 2 \\
	\textbf{} & \textbf{PART} & 0 & 1 & 2 & 0 & 0 & 0 & 0 & 0 & 0 & 0 & 0 & 0 & 0 & 0 & 0 & 0 \\
	\textbf{} & \textbf{PRON} & 0 & 1 & 1 & 0 & 0 & 0 & 16 & 0 & 9 & 0 & 409 & 1 & 0 & 0 & 1 & 0 \\
	\textbf{} & \textbf{PROPN} & 70 & 4 & 0 & 0 & 0 & 0 & 380 & 4 & 11 & 0 & 4 & 1006 & 0 & 0 & 1 & 21 \\
	\textbf{} & \textbf{SCONJ} & 0 & 62 & 34 & 0 & 6 & 1 & 1 & 0 & 0 & 0 & 0 & 0 & 2 & 0 & 0 & 0 \\
	\textbf{} & \textbf{SYM} & 0 & 0 & 0 & 0 & 0 & 0 & 9 & 0 & 0 & 0 & 0 & 0 & 0 & 26 & 0 & 0 \\
	\textbf{} & \textbf{VERB} & 63 & 19 & 2 & 36 & 0 & 0 & 100 & 0 & 49 & 0 & 1 & 1 & 0 & 0 & 1508 & 0 \\
	\textbf{} & \textbf{X} & 1 & 0 & 0 & 0 & 0 & 0 & 3 & 0 & 0 & 0 & 0 & 1 & 0 & 0 & 0 & 6 \\
	\textbf{En\textbackslash Zh} & \textbf{} &  &  &  &  &  &  &  &  &  &  &  &  &  &  &  &  \\
	\textbf{} & \textbf{ADJ} & 337 & 6 & 30 & 9 & 0 & 42 & 373 & 66 & 13 & 0 & 6 & 131 & 0 & 0 & 98 & 2 \\
	\textbf{} & \textbf{ADP} & 5 & 227 & 3 & 0 & 6 & 2 & 18 & 0 & 34 & 2 & 0 & 0 & 0 & 1 & 117 & 0 \\
	\textbf{} & \textbf{ADV} & 46 & 20 & 338 & 9 & 1 & 2 & 79 & 3 & 33 & 0 & 12 & 0 & 0 & 2 & 42 & 1 \\
	\textbf{} & \textbf{AUX} & 0 & 0 & 2 & 50 & 0 & 0 & 2 & 0 & 6 & 0 & 0 & 0 & 0 & 0 & 11 & 0 \\
	\textbf{} & \textbf{CCONJ} & 0 & 4 & 135 & 2 & 243 & 0 & 1 & 0 & 51 & 0 & 0 & 0 & 0 & 2 & 1 & 1 \\
	\textbf{} & \textbf{DET} & 7 & 0 & 7 & 4 & 0 & 36 & 4 & 6 & 1 & 0 & 3 & 0 & 0 & 0 & 9 & 0 \\
	\textbf{} & \textbf{INTJ} & 0 & 0 & 0 & 0 & 0 & 0 & 0 & 0 & 0 & 0 & 0 & 0 & 0 & 0 & 0 & 1 \\
	\textbf{} & \textbf{NOUN} & 54 & 7 & 4 & 3 & 0 & 7 & 2671 & 16 & 38 & 4 & 4 & 5 & 0 & 0 & 356 & 10 \\
	\textbf{} & \textbf{NUM} & 2 & 0 & 0 & 0 & 0 & 2 & 2 & 194 & 1 & 0 & 0 & 9 & 0 & 0 & 0 & 0 \\
	\textbf{} & \textbf{None} & 49 & 130 & 462 & 48 & 15 & 21 & 559 & 31 & 0 & 11 & 100 & 39 & 1 & 4 & 464 & 23 \\
	\textbf{} & \textbf{PART} & 0 & 4 & 22 & 9 & 0 & 0 & 0 & 0 & 0 & 0 & 0 & 0 & 0 & 0 & 13 & 0 \\
	\textbf{} & \textbf{PRON} & 1 & 0 & 1 & 0 & 0 & 7 & 29 & 0 & 54 & 0 & 526 & 5 & 0 & 0 & 2 & 2 \\
	\textbf{} & \textbf{PROPN} & 6 & 0 & 0 & 0 & 0 & 0 & 192 & 2 & 7 & 2 & 5 & 1001 & 0 & 0 & 11 & 126 \\
	\textbf{} & \textbf{SCONJ} & 1 & 56 & 7 & 2 & 0 & 0 & 3 & 0 & 6 & 0 & 0 & 0 & 0 & 8 & 14 & 1 \\
	\textbf{} & \textbf{SYM} & 0 & 0 & 0 & 0 & 1 & 0 & 2 & 0 & 1 & 0 & 0 & 0 & 0 & 0 & 0 & 0 \\
	\textbf{} & \textbf{VERB} & 32 & 14 & 25 & 25 & 0 & 2 & 110 & 0 & 54 & 1 & 0 & 0 & 1 & 0 & 1394 & 2 \\
	\textbf{} & \textbf{X} & 1 & 0 & 0 & 0 & 0 & 0 & 3 & 0 & 0 & 0 & 1 & 0 & 0 & 0 & 0 & 5 \\
	\textbf{En\textbackslash Ko} & \textbf{} &  &  &  &  &  &  &  &  &  &  &  &  &  &  &  &  \\
	\textbf{} & \textbf{ADJ} & 152 & 0 & 37 & 0 & 0 & 42 & 554 & 20 & 14 & 2 & 4 & 128 & 0 & 0 & 31 & 0 \\
	\textbf{} & \textbf{ADP} & 10 & 0 & 13 & 0 & 0 & 0 & 100 & 0 & 68 & 6 & 1 & 0 & 0 & 0 & 27 & 0 \\
	\textbf{} & \textbf{ADV} & 46 & 0 & 202 & 1 & 17 & 6 & 125 & 1 & 28 & 8 & 5 & 2 & 0 & 0 & 21 & 0 \\
	\textbf{} & \textbf{AUX} & 7 & 0 & 0 & 23 & 0 & 0 & 6 & 0 & 9 & 1 & 0 & 0 & 0 & 0 & 11 & 0 \\
	\textbf{} & \textbf{CCONJ} & 2 & 0 & 0 & 0 & 19 & 0 & 2 & 0 & 0 & 0 & 0 & 1 & 0 & 0 & 1 & 0 \\
	\textbf{} & \textbf{DET} & 5 & 0 & 7 & 0 & 0 & 49 & 13 & 0 & 5 & 2 & 2 & 0 & 0 & 0 & 5 & 0 \\
	\textbf{} & \textbf{INTJ} & 0 & 0 & 0 & 0 & 0 & 0 & 0 & 0 & 0 & 0 & 0 & 0 & 0 & 0 & 0 & 1 \\
	\textbf{} & \textbf{NOUN} & 10 & 0 & 13 & 6 & 0 & 8 & 2628 & 4 & 39 & 2 & 8 & 29 & 0 & 0 & 46 & 0 \\
	\textbf{} & \textbf{NUM} & 0 & 0 & 1 & 0 & 0 & 45 & 21 & 147 & 1 & 0 & 0 & 1 & 0 & 0 & 0 & 0 \\
	\textbf{} & \textbf{None} & 0 & 0 & 0 & 0 & 0 & 0 & 3 & 1 & 0 & 0 & 0 & 0 & 0 & 0 & 0 & 0 \\
	\textbf{} & \textbf{PART} & 11 & 0 & 2 & 0 & 0 & 0 & 0 & 0 & 2 & 2 & 0 & 0 & 0 & 0 & 26 & 0 \\
	\textbf{} & \textbf{PRON} & 3 & 0 & 0 & 1 & 0 & 18 & 17 & 0 & 50 & 0 & 232 & 6 & 0 & 0 & 0 & 0 \\
	\textbf{} & \textbf{PROPN} & 0 & 0 & 0 & 1 & 0 & 0 & 181 & 7 & 10 & 0 & 0 & 1002 & 0 & 0 & 0 & 0 \\
	\textbf{} & \textbf{SCONJ} & 0 & 0 & 1 & 0 & 0 & 0 & 14 & 0 & 0 & 1 & 0 & 1 & 0 & 0 & 9 & 0 \\
	\textbf{} & \textbf{SYM} & 0 & 0 & 0 & 0 & 0 & 0 & 27 & 0 & 0 & 0 & 0 & 0 & 0 & 0 & 0 & 0 \\
	\textbf{} & \textbf{VERB} & 32 & 0 & 12 & 5 & 0 & 1 & 804 & 0 & 39 & 30 & 0 & 5 & 0 & 0 & 400 & 0 \\
	\textbf{} & \textbf{X} & 0 & 0 & 0 & 0 & 0 & 0 & 4 & 0 & 0 & 1 & 0 & 2 & 0 & 0 & 0 & 0 \\
	\textbf{En\textbackslash Jp} & \textbf{} &  &  &  &  &  &  &  &  &  &  &  &  &  &  &  &  \\
	\textbf{} & \textbf{ADJ} &   214 &    13 &   28 &      5 &    0 &     0 &   560 &   20 &    61 &     0 &     2 &    101 &      0 &    0 &    32 &  0 \\
	\textbf{} & \textbf{ADP} &     3 &  1106 &   19 &      7 &    0 &     0 &   127 &    1 &    27 &     3 &     1 &     14 &      6 &    0 &    39 &  0 \\
	\textbf{} & \textbf{ADV} &    47 &    49 &  140 &     36 &    0 &     0 &    89 &    0 &   113 &    19 &     4 &      4 &      8 &    0 &    19 &  0 \\
	\textbf{} & \textbf{CCONJ} &     2 &   187 &    4 &     73 &    0 &     0 &     3 &    0 &     0 &     3 &     0 &      1 &     50 &    1 &    16 &  0 \\
	\textbf{} & \textbf{DET} &    15 &   154 &   12 &      4 &   99 &     0 &    55 &    1 &    12 &     1 &    13 &      8 &      1 &    0 &     4 &  0 \\
	\textbf{} & \textbf{INTJ} &     0 &     0 &    0 &      0 &    0 &     0 &     1 &    0 &     0 &     0 &     0 &      0 &      0 &    0 &     0 &  0 \\
	\textbf{} & \textbf{NOUN} &    22 &     4 &   14 &      5 &    0 &     0 &  3180 &    0 &   174 &    12 &     3 &     89 &      2 &    0 &    58 &  0 \\
	\textbf{} & \textbf{NUM} &     1 &     0 &    0 &      0 &    0 &     0 &    10 &  125 &    22 &     0 &     0 &      1 &      0 &    0 &     3 &  0 \\
	\textbf{} & \textbf{None} &     0 &     0 &    0 &      0 &    0 &     0 &     0 &    0 &     0 &     0 &     0 &      0 &      0 &    0 &     1 &  0 \\
	\textbf{} & \textbf{PART} &     2 &    99 &    2 &      0 &    0 &     0 &     5 &    0 &     5 &     0 &     0 &      1 &     14 &    0 &     2 &  0 \\
	\textbf{} & \textbf{PRON} &     1 &    12 &    3 &      2 &   17 &     0 &    31 &    0 &   464 &     1 &   233 &      1 &      8 &    0 &     7 &  0 \\
	\textbf{} & \textbf{PROPN} &     2 &     0 &    0 &      0 &    0 &     0 &   392 &    0 &    33 &     0 &     0 &   1223 &      0 &    0 &     0 &  0 \\
	\textbf{} & \textbf{SCONJ} &     1 &    39 &    4 &      1 &    0 &     0 &    21 &    0 &     2 &     7 &     0 &      0 &     45 &    0 &     1 &  0 \\
	\textbf{} & \textbf{SYM} &     0 &     0 &    0 &      0 &    0 &     0 &    30 &    3 &     6 &     0 &     0 &      0 &      0 &    0 &     0 &  0 \\
	\textbf{} & \textbf{VERB} &    14 &    13 &    2 &      0 &    0 &     0 &   145 &    0 &   145 &     1 &     0 &      3 &      1 &    0 &   497 &  0 \\
	\textbf{} & \textbf{X} &     0 &     0 &    0 &      0 &    0 &     0 &     2 &    0 &     3 &     0 &     0 &      6 &      0 &    0 &     0 &  0 \\
\end{tabular}
}
\caption{Raw counts of translation mappings of parts of speech.}
\label{table:ud-divergences-pos-raw}
\end{table*}

\begin{table*}[t]
	\centering
	\scalebox{.55}{%
	\begin{tabular}{llllllllllllllllll}
	\textbf{En\textbackslash Ru} & \textbf{} & \textbf{ADJ} & \textbf{ADP} & \textbf{ADV} & \textbf{AUX} & \textbf{CCONJ} & \textbf{DET} & \textbf{NOUN} & \textbf{NUM} & \textbf{None} & \textbf{PART} & \textbf{PRON} & \textbf{PROPN} & \textbf{SCONJ} & \textbf{SYM} & \textbf{VERB} & \textbf{X} \\
	\textbf{} & \textbf{ADJ} & 79 & 0 & 4 & 0 & 0 & 2 & 6 & 0 & 2 & 0 & 0 & 2 & 0 & 0 & 4 & 1 \\
	\textbf{} & \textbf{ADP} & 1 & 86 & 3 & 0 & 0 & 1 & 3 & 0 & 3 & 0 & 0 & 0 & 0 & 0 & 2 & 1 \\
	\textbf{} & \textbf{ADV} & 6 & 2 & 68 & 0 & 5 & 1 & 3 & 0 & 6 & 5 & 0 & 0 & 0 & 0 & 1 & 2 \\
	\textbf{} & \textbf{AUX} & 7 & 0 & 2 & 5 & 0 & 0 & 1 & 0 & 3 & 0 & 0 & 0 & 0 & 0 & 82 & 0 \\
	\textbf{} & \textbf{CCONJ} & 0 & 0 & 0 & 0 & 93 & 0 & 0 & 0 & 7 & 0 & 0 & 0 & 0 & 0 & 0 & 0 \\
	\textbf{} & \textbf{DET} & 34 & 1 & 3 & 0 & 0 & 32 & 3 & 10 & 4 & 9 & 1 & 0 & 0 & 0 & 1 & 0 \\
	\textbf{} & \textbf{INTJ} & 0 & 0 & 0 & 0 & 0 & 0 & 0 & 0 & 0 & 0 & 0 & 0 & 0 & 0 & 0 & 100 \\
	\textbf{} & \textbf{NOUN} & 6 & 0 & 1 & 0 & 0 & 0 & 86 & 0 & 2 & 0 & 0 & 3 & 0 & 0 & 1 & 0 \\
	\textbf{} & \textbf{NUM} & 7 & 0 & 0 & 0 & 0 & 0 & 8 & 81 & 1 & 0 & 0 & 0 & 0 & 0 & 0 & 1 \\
	\textbf{} & \textbf{None} & 10 & 4 & 9 & 0 & 1 & 4 & 39 & 1 & 0 & 3 & 6 & 3 & 0 & 0 & 18 & 1 \\
	\textbf{} & \textbf{PART} & 0 & 2 & 2 & 0 & 0 & 0 & 0 & 0 & 0 & 90 & 0 & 0 & 0 & 0 & 6 & 0 \\
	\textbf{} & \textbf{PRON} & 1 & 0 & 0 & 0 & 0 & 24 & 3 & 0 & 4 & 0 & 67 & 0 & 0 & 0 & 0 & 0 \\
	\textbf{} & \textbf{PROPN} & 9 & 0 & 0 & 0 & 0 & 0 & 8 & 0 & 1 & 0 & 0 & 81 & 0 & 0 & 0 & 1 \\
	\textbf{} & \textbf{SCONJ} & 0 & 68 & 16 & 0 & 0 & 0 & 0 & 0 & 0 & 16 & 0 & 0 & 0 & 0 & 0 & 0 \\
	\textbf{} & \textbf{SYM} & 0 & 0 & 0 & 0 & 0 & 0 & 12 & 0 & 0 & 0 & 0 & 0 & 0 & 88 & 0 & 0 \\
	\textbf{} & \textbf{VERB} & 3 & 0 & 1 & 1 & 0 & 0 & 8 & 0 & 4 & 0 & 0 & 0 & 0 & 0 & 83 & 0 \\
	\textbf{} & \textbf{X} & 20 & 0 & 0 & 0 & 0 & 0 & 20 & 0 & 0 & 0 & 0 & 60 & 0 & 0 & 0 & 0 \\
	\textbf{En\textbackslash Fr} & \textbf{} &  &  &  &  &  &  &  &  &  &  &  &  &  &  &  &  \\
	\textbf{} & \textbf{ADJ} & 77 & 1 & 4 & 0 & 0 & 0 & 10 & 0 & 3 & 0 & 0 & 1 & 0 & 0 & 3 & 0 \\
	\textbf{} & \textbf{ADP} & 1 & 86 & 5 & 0 & 0 & 1 & 3 & 0 & 2 & 0 & 0 & 0 & 0 & 0 & 2 & 0 \\
	\textbf{} & \textbf{ADV} & 5 & 5 & 72 & 0 & 1 & 0 & 6 & 0 & 7 & 0 & 1 & 0 & 1 & 0 & 1 & 0 \\
	\textbf{} & \textbf{AUX} & 0 & 0 & 2 & 5 & 0 & 0 & 0 & 0 & 2 & 0 & 0 & 0 & 0 & 0 & 90 & 0 \\
	\textbf{} & \textbf{CCONJ} & 0 & 0 & 2 & 0 & 92 & 0 & 0 & 0 & 6 & 0 & 0 & 0 & 0 & 0 & 0 & 0 \\
	\textbf{} & \textbf{DET} & 38 & 3 & 6 & 0 & 0 & 29 & 6 & 8 & 8 & 0 & 3 & 0 & 0 & 0 & 0 & 0 \\
	\textbf{} & \textbf{INTJ} & 0 & 0 & 0 & 0 & 0 & 0 & 0 & 0 & 0 & 0 & 0 & 0 & 0 & 0 & 100 & 0 \\
	\textbf{} & \textbf{NOUN} & 3 & 0 & 0 & 0 & 0 & 0 & 93 & 0 & 1 & 0 & 0 & 0 & 0 & 0 & 1 & 0 \\
	\textbf{} & \textbf{NUM} & 2 & 0 & 0 & 0 & 0 & 3 & 4 & 90 & 0 & 0 & 0 & 0 & 0 & 0 & 0 & 0 \\
	\textbf{} & \textbf{None} & 9 & 5 & 11 & 0 & 9 & 1 & 27 & 0 & 0 & 0 & 9 & 0 & 0 & 0 & 28 & 0 \\
	\textbf{} & \textbf{PART} & 0 & 33 & 67 & 0 & 0 & 0 & 0 & 0 & 0 & 0 & 0 & 0 & 0 & 0 & 0 & 0 \\
	\textbf{} & \textbf{PRON} & 0 & 0 & 0 & 0 & 0 & 0 & 4 & 0 & 2 & 0 & 93 & 0 & 0 & 0 & 0 & 0 \\
	\textbf{} & \textbf{PROPN} & 5 & 0 & 0 & 0 & 0 & 0 & 25 & 0 & 1 & 0 & 0 & 67 & 0 & 0 & 0 & 1 \\
	\textbf{} & \textbf{SCONJ} & 0 & 58 & 32 & 0 & 6 & 1 & 1 & 0 & 0 & 0 & 0 & 0 & 2 & 0 & 0 & 0 \\
	\textbf{} & \textbf{SYM} & 0 & 0 & 0 & 0 & 0 & 0 & 26 & 0 & 0 & 0 & 0 & 0 & 0 & 74 & 0 & 0 \\
	\textbf{} & \textbf{VERB} & 4 & 1 & 0 & 2 & 0 & 0 & 6 & 0 & 3 & 0 & 0 & 0 & 0 & 0 & 85 & 0 \\
	\textbf{} & \textbf{X} & 9 & 0 & 0 & 0 & 0 & 0 & 27 & 0 & 0 & 0 & 0 & 9 & 0 & 0 & 0 & 55 \\
	\textbf{En-Zh} & \textbf{} &  &  &  &  &  &  &  &  &  &  &  &  &  &  &  &  \\
	\textbf{} & \textbf{ADJ} & 30 & 1 & 3 & 1 & 0 & 4 & 34 & 6 & 1 & 0 & 1 & 12 & 0 & 0 & 9 & 0 \\
	\textbf{} & \textbf{ADP} & 1 & 55 & 1 & 0 & 1 & 0 & 4 & 0 & 8 & 0 & 0 & 0 & 0 & 0 & 28 & 0 \\
	\textbf{} & \textbf{ADV} & 8 & 3 & 57 & 2 & 0 & 0 & 13 & 1 & 6 & 0 & 2 & 0 & 0 & 0 & 7 & 0 \\
	\textbf{} & \textbf{AUX} & 0 & 0 & 3 & 70 & 0 & 0 & 3 & 0 & 8 & 0 & 0 & 0 & 0 & 0 & 15 & 0 \\
	\textbf{} & \textbf{CCONJ} & 0 & 1 & 31 & 0 & 55 & 0 & 0 & 0 & 12 & 0 & 0 & 0 & 0 & 0 & 0 & 0 \\
	\textbf{} & \textbf{DET} & 9 & 0 & 9 & 5 & 0 & 47 & 5 & 8 & 1 & 0 & 4 & 0 & 0 & 0 & 12 & 0 \\
	\textbf{} & \textbf{INTJ} & 0 & 0 & 0 & 0 & 0 & 0 & 0 & 0 & 0 & 0 & 0 & 0 & 0 & 0 & 0 & 100 \\
	\textbf{} & \textbf{NOUN} & 2 & 0 & 0 & 0 & 0 & 0 & 84 & 1 & 1 & 0 & 0 & 0 & 0 & 0 & 11 & 0 \\
	\textbf{} & \textbf{NUM} & 1 & 0 & 0 & 0 & 0 & 1 & 1 & 92 & 0 & 0 & 0 & 4 & 0 & 0 & 0 & 0 \\
	\textbf{} & \textbf{None} & 3 & 7 & 24 & 2 & 1 & 1 & 29 & 2 & 0 & 1 & 5 & 2 & 0 & 0 & 24 & 1 \\
	\textbf{} & \textbf{PART} & 0 & 8 & 46 & 19 & 0 & 0 & 0 & 0 & 0 & 0 & 0 & 0 & 0 & 0 & 27 & 0 \\
	\textbf{} & \textbf{PRON} & 0 & 0 & 0 & 0 & 0 & 1 & 5 & 0 & 9 & 0 & 84 & 1 & 0 & 0 & 0 & 0 \\
	\textbf{} & \textbf{PROPN} & 0 & 0 & 0 & 0 & 0 & 0 & 14 & 0 & 1 & 0 & 0 & 74 & 0 & 0 & 1 & 9 \\
	\textbf{} & \textbf{SCONJ} & 1 & 57 & 7 & 2 & 0 & 0 & 3 & 0 & 6 & 0 & 0 & 0 & 0 & 8 & 14 & 1 \\
	\textbf{} & \textbf{SYM} & 0 & 0 & 0 & 0 & 25 & 0 & 50 & 0 & 25 & 0 & 0 & 0 & 0 & 0 & 0 & 0 \\
	\textbf{} & \textbf{VERB} & 2 & 1 & 2 & 2 & 0 & 0 & 7 & 0 & 3 & 0 & 0 & 0 & 0 & 0 & 84 & 0 \\
	\textbf{} & \textbf{X} & 10 & 0 & 0 & 0 & 0 & 0 & 30 & 0 & 0 & 0 & 10 & 0 & 0 & 0 & 0 & 50 \\
	\textbf{En\textbackslash Ko} & \textbf{} &  &  &  &  &  &  &  &  &  &  &  &  &  &  &  &  \\
	\textbf{} & \textbf{ADJ} & 15 & 0 & 4 & 0 & 0 & 4 & 56 & 2 & 1 & 0 & 0 & 13 & 0 & 0 & 3 & 0 \\
	\textbf{} & \textbf{ADP} & 4 & 0 & 6 & 0 & 0 & 0 & 44 & 0 & 30 & 3 & 0 & 0 & 0 & 0 & 12 & 0 \\
	\textbf{} & \textbf{ADV} & 10 & 0 & 44 & 0 & 4 & 1 & 27 & 0 & 6 & 2 & 1 & 0 & 0 & 0 & 5 & 0 \\
	\textbf{} & \textbf{AUX} & 12 & 0 & 0 & 40 & 0 & 0 & 11 & 0 & 16 & 2 & 0 & 0 & 0 & 0 & 19 & 0 \\
	\textbf{} & \textbf{CCONJ} & 8 & 0 & 0 & 0 & 76 & 0 & 8 & 0 & 0 & 0 & 0 & 4 & 0 & 0 & 4 & 0 \\
	\textbf{} & \textbf{DET} & 6 & 0 & 8 & 0 & 0 & 56 & 15 & 0 & 6 & 2 & 2 & 0 & 0 & 0 & 6 & 0 \\
	\textbf{} & \textbf{INTJ} & 0 & 0 & 0 & 0 & 0 & 0 & 0 & 0 & 0 & 0 & 0 & 0 & 0 & 0 & 0 & 100 \\
	\textbf{} & \textbf{NOUN} & 0 & 0 & 0 & 0 & 0 & 0 & 94 & 0 & 1 & 0 & 0 & 1 & 0 & 0 & 2 & 0 \\
	\textbf{} & \textbf{NUM} & 0 & 0 & 0 & 0 & 0 & 21 & 10 & 68 & 0 & 0 & 0 & 0 & 0 & 0 & 0 & 0 \\
	\textbf{} & \textbf{None} & 0 & 0 & 0 & 0 & 0 & 0 & 75 & 25 & 0 & 0 & 0 & 0 & 0 & 0 & 0 & 0 \\
	\textbf{} & \textbf{PART} & 26 & 0 & 5 & 0 & 0 & 0 & 0 & 0 & 5 & 5 & 0 & 0 & 0 & 0 & 60 & 0 \\
	\textbf{} & \textbf{PRON} & 1 & 0 & 0 & 0 & 0 & 6 & 5 & 0 & 15 & 0 & 71 & 2 & 0 & 0 & 0 & 0 \\
	\textbf{} & \textbf{PROPN} & 0 & 0 & 0 & 0 & 0 & 0 & 15 & 1 & 1 & 0 & 0 & 83 & 0 & 0 & 0 & 0 \\
	\textbf{} & \textbf{SCONJ} & 0 & 0 & 4 & 0 & 0 & 0 & 54 & 0 & 0 & 4 & 0 & 4 & 0 & 0 & 35 & 0 \\
	\textbf{} & \textbf{SYM} & 0 & 0 & 0 & 0 & 0 & 0 & 100 & 0 & 0 & 0 & 0 & 0 & 0 & 0 & 0 & 0 \\
	\textbf{} & \textbf{VERB} & 2 & 0 & 1 & 0 & 0 & 0 & 61 & 0 & 3 & 2 & 0 & 0 & 0 & 0 & 30 & 0 \\
	\textbf{} & \textbf{X} & 0 & 0 & 0 & 0 & 0 & 0 & 57 & 0 & 0 & 14 & 0 & 29 & 0 & 0 & 0 & 0 \\
	\textbf{En\textbackslash Jp} & \textbf{} &  &  &  &  &  &  &  &  &  &  &  &  &  &  &  &  \\
	\textbf{} & \textbf{ADJ} &    20 &    1 &    3 &      0 &    0 &     0 &    54 &    2 &     6 &     0 &     0 &     10 &      0 &    0 &     3 &  0 \\
	\textbf{} & \textbf{ADP} &     0 &   81 &    1 &      0 &    0 &     0 &     9 &    0 &     2 &     0 &     0 &      1 &      0 &    0 &     3 &  0 \\
	\textbf{} & \textbf{ADV} &     9 &    9 &   26 &      7 &    0 &     0 &    16 &    0 &    21 &     4 &     1 &      1 &      2 &    0 &     4 &  0 \\
	\textbf{} & \textbf{CCONJ} &     0 &   47 &    1 &     18 &    0 &     0 &     1 &    0 &     0 &     1 &     0 &      0 &     12 &    0 &     4 &  0 \\
	\textbf{} & \textbf{DET} &     4 &   40 &    3 &      1 &   26 &     0 &    14 &    0 &     3 &     0 &     3 &      2 &      0 &    0 &     1 &  0 \\
	\textbf{} & \textbf{INTJ} &     0 &    0 &    0 &      0 &    0 &     0 &   100 &    0 &     0 &     0 &     0 &      0 &      0 &    0 &     0 &  0 \\
	\textbf{} & \textbf{NOUN} &     1 &    0 &    0 &      0 &    0 &     0 &    89 &    0 &     5 &     0 &     0 &      2 &      0 &    0 &     2 &  0 \\
	\textbf{} & \textbf{NUM} &     1 &    0 &    0 &      0 &    0 &     0 &     6 &   77 &    14 &     0 &     0 &      1 &      0 &    0 &     2 &  0 \\
	\textbf{} & \textbf{None} &     0 &    0 &    0 &      0 &    0 &     0 &     0 &    0 &     0 &     0 &     0 &      0 &      0 &    0 &    50 &  0 \\
	\textbf{} & \textbf{PART} &     1 &   61 &    1 &      0 &    0 &     0 &     3 &    0 &     3 &     0 &     0 &      1 &      9 &    0 &     1 &  0 \\
	\textbf{} & \textbf{PRON} &     0 &    2 &    0 &      0 &    2 &     0 &     4 &    0 &    59 &     0 &    30 &      0 &      1 &    0 &     1 &  0 \\
	\textbf{} & \textbf{PROPN} &     0 &    0 &    0 &      0 &    0 &     0 &    24 &    0 &     2 &     0 &     0 &     74 &      0 &    0 &     0 &  0 \\
	\textbf{} & \textbf{SCONJ} &     1 &   30 &    3 &      1 &    0 &     0 &    16 &    0 &     2 &     5 &     0 &      0 &     35 &    0 &     1 &  0 \\
	\textbf{} & \textbf{SYM} &     0 &    0 &    0 &      0 &    0 &     0 &    77 &    8 &    15 &     0 &     0 &      0 &      0 &    0 &     0 &  0 \\
	\textbf{} & \textbf{VERB} &     2 &    2 &    0 &      0 &    0 &     0 &    17 &    0 &    17 &     0 &     0 &      0 &      0 &    0 &    58 &  0 \\
	\textbf{} & \textbf{X} &     0 &    0 &    0 &      0 &    0 &     0 &    18 &    0 &    27 &     0 &     0 &     54 &      0 &    0 &     0 &  0 \\
	\end{tabular}
	}
	\caption{Percentages of translation mappings of parts of speech.}
\label{table:ud-divergences-pos-percent}
\end{table*}

\subsection{UD edge types}\label{sssec:appendix:edges}

Raw-count confusion matrices for translation equivalents of major UD edge types in three parallel corpora are presetned in Table~\ref{table:ud-divergences-edges-raw}.

\begin{table*}[t]
\centering
\scalebox{0.55}{%
\begin{tabular}{lllllllllllllllllllllll}
	\textbf{En\textbackslash Ru} & & \rotatebox{90}{\textbf{acl}} & 
	\rotatebox{90}{\textbf{advcl}} & 
	\rotatebox{90}{\textbf{advmod}} & 
	\rotatebox{90}{\textbf{amod}} & 
	\rotatebox{90}{\textbf{appos}} & 
	\rotatebox{90}{\textbf{ccomp}} & 
	\rotatebox{90}{\textbf{compound}} & 
	\rotatebox{90}{\textbf{conj}} & 
	\rotatebox{90}{\textbf{fixed}} & 
	\rotatebox{90}{\textbf{flat}} & 
	\rotatebox{90}{\textbf{nmod}} & 
	\rotatebox{90}{\textbf{nsubj}} & 
	\rotatebox{90}{\textbf{nummod}} & 
	\rotatebox{90}{\textbf{obj}} & 
	\rotatebox{90}{\textbf{obl}} & 
	\rotatebox{90}{\textbf{parataxis}} & 
	\rotatebox{90}{\textbf{xcomp}} & 
	\rotatebox{90}{\textbf{Collapsed}} & 
	\rotatebox{90}{\textbf{Other}} & 
	\rotatebox{90}{\textbf{MCOP \#}} & 
	\rotatebox{90}{\textbf{MCOP}} \\
	\textbf{} & \textbf{acl} & 151 & 2 & 1 & 5 & 1 & 2 & 0 & 1 & 0 & 0 & 27 & 7 & 0 & 5 & 3 & 3 & 13 & 2 & 89 & 15 & nmod+acl \\
	\textbf{} & \textbf{advcl} & 11 & 68 & 2 & 0 & 0 & 9 & 0 & 11 & 0 & 0 & 2 & 0 & 0 & 1 & 22 & 4 & 2 & 1 & 79 & 5 & advcl+xcomp \\
	\textbf{} & \textbf{advmod} & 0 & 0 & 332 & 11 & 0 & 0 & 0 & 1 & 0 & 1 & 4 & 0 & 1 & 1 & 16 & 2 & 2 & 27 & 129 & 11 & advmod+nummod \\
	\textbf{} & \textbf{amod} & 29 & 0 & 8 & 893 & 4 & 4 & 0 & 1 & 0 & 5 & 60 & 4 & 7 & 1 & 3 & 0 & 0 & 57 & 77 & 18 & det \\
	\textbf{} & \textbf{appos} & 0 & 0 & 0 & 0 & 43 & 0 & 0 & 1 & 0 & 15 & 21 & 1 & 1 & 0 & 0 & 1 & 0 & 1 & 35 & 7 & nmod+flat \\
	\textbf{} & \textbf{ccomp} & 1 & 2 & 1 & 0 & 0 & 48 & 0 & 0 & 0 & 1 & 0 & 0 & 0 & 1 & 0 & 6 & 6 & 0 & 32 & 11 & ccomp+xcomp \\
	\textbf{} & \textbf{compound} & 0 & 0 & 2 & 244 & 7 & 0 & 4 & 3 & 0 & 48 & 173 & 0 & 10 & 2 & 4 & 0 & 0 & 153 & 97 & 22 & nmod+nmod \\
	\textbf{} & \textbf{conj} & 0 & 2 & 0 & 1 & 0 & 0 & 0 & 387 & 1 & 1 & 0 & 0 & 1 & 0 & 1 & 1 & 0 & 5 & 108 & 10 & conj+conj \\
	\textbf{} & \textbf{fixed} & 0 & 0 & 0 & 0 & 0 & 0 & 0 & 0 & 0 & 2 & 1 & 0 & 0 & 0 & 0 & 0 & 0 & 9 & 1 & 1 & nmod+flat \\
	\textbf{} & \textbf{flat} & 0 & 0 & 0 & 6 & 3 & 0 & 2 & 0 & 0 & 154 & 13 & 0 & 0 & 0 & 1 & 0 & 0 & 8 & 19 & 7 & flat+flat \\
	\textbf{} & \textbf{nmod} & 4 & 1 & 11 & 40 & 6 & 1 & 0 & 5 & 0 & 10 & 632 & 5 & 7 & 6 & 31 & 0 & 0 & 5 & 309 & 67 & det \\
	\textbf{} & \textbf{nsubj} & 1 & 1 & 1 & 1 & 0 & 0 & 0 & 1 & 0 & 0 & 14 & 829 & 0 & 22 & 26 & 0 & 0 & 3 & 261 & 48 & nsubj+xcomp \\
	\textbf{} & \textbf{nummod} & 0 & 0 & 1 & 10 & 0 & 0 & 0 & 0 & 0 & 1 & 3 & 0 & 153 & 0 & 1 & 0 & 0 & 14 & 4 & 1 & compound+nummod \\
	\textbf{} & \textbf{obj} & 1 & 0 & 4 & 4 & 0 & 1 & 0 & 0 & 0 & 0 & 58 & 27 & 0 & 393 & 62 & 0 & 3 & 25 & 106 & 26 & iobj \\
	\textbf{} & \textbf{obl} & 0 & 5 & 44 & 2 & 0 & 0 & 0 & 3 & 0 & 0 & 38 & 17 & 0 & 27 & 463 & 3 & 2 & 10 & 311 & 67 & iobj \\
	\textbf{} & \textbf{parataxis} & 0 & 1 & 0 & 0 & 0 & 1 & 0 & 5 & 0 & 0 & 0 & 0 & 0 & 0 & 0 & 41 & 0 & 0 & 22 & 2 & acl+obl \\
	\textbf{} & \textbf{xcomp} & 0 & 5 & 9 & 0 & 0 & 7 & 0 & 2 & 0 & 0 & 1 & 3 & 0 & 3 & 15 & 0 & 91 & 4 & 40 & 15 & iobj \\
	\textbf{En\textbackslash Fr} & \textbf{} &  &  &  &  &  &  &  &  &  &  &  &  &  &  &  &  &  &  &  &  &  \\
	\textbf{} & \textbf{acl} & 104 & 4 & 1 & 13 & 3 & 52 & 0 & 3 & 0 & 0 & 12 & 1 & 0 & 4 & 3 & 0 & 24 & 2 & 80 & 8 & acl+xcomp \\
	\textbf{} & \textbf{advcl} & 2 & 96 & 2 & 0 & 0 & 19 & 0 & 4 & 0 & 0 & 0 & 0 & 0 & 0 & 16 & 4 & 5 & 0 & 70 & 9 & xcomp+advcl \\
	\textbf{} & \textbf{advmod} & 1 & 0 & 321 & 7 & 2 & 0 & 0 & 1 & 2 & 0 & 8 & 1 & 0 & 6 & 17 & 0 & 6 & 31 & 108 & 18 & advmod+xcomp \\
	\textbf{} & \textbf{amod} & 1 & 0 & 9 & 854 & 3 & 14 & 5 & 2 & 0 & 0 & 97 & 3 & 3 & 1 & 5 & 0 & 0 & 61 & 80 & 16 & det \\
	\textbf{} & \textbf{appos} & 0 & 0 & 0 & 2 & 88 & 1 & 0 & 1 & 0 & 0 & 7 & 0 & 0 & 0 & 1 & 0 & 0 & 0 & 27 & 4 & appos+nmod \\
	\textbf{} & \textbf{ccomp} & 0 & 1 & 0 & 0 & 0 & 51 & 0 & 0 & 0 & 0 & 0 & 0 & 0 & 2 & 0 & 5 & 11 & 0 & 28 & 10 & ccomp+xcomp \\
	\textbf{} & \textbf{compound} & 0 & 0 & 1 & 143 & 32 & 0 & 47 & 5 & 0 & 21 & 300 & 1 & 12 & 4 & 5 & 0 & 2 & 102 & 46 & 10 & nmod+nmod \\
	\textbf{} & \textbf{conj} & 1 & 2 & 0 & 0 & 2 & 2 & 0 & 405 & 0 & 0 & 3 & 0 & 0 & 1 & 0 & 6 & 1 & 7 & 96 & 9 & nmod+conj \\
	\textbf{} & \textbf{fixed} & 0 & 0 & 0 & 1 & 1 & 0 & 0 & 0 & 0 & 1 & 0 & 0 & 0 & 0 & 0 & 0 & 0 & 1 & 2 & 1 & appos+nummod \\
	\textbf{} & \textbf{flat} & 0 & 0 & 0 & 5 & 32 & 0 & 5 & 0 & 0 & 140 & 4 & 0 & 8 & 0 & 0 & 0 & 0 & 8 & 15 & 7 & appos+flat \\
	\textbf{} & \textbf{nmod} & 0 & 0 & 6 & 17 & 3 & 0 & 1 & 2 & 0 & 1 & 702 & 4 & 12 & 13 & 34 & 0 & 1 & 15 & 253 & 37 & nmod+nmod \\
	\textbf{} & \textbf{nsubj} & 4 & 0 & 1 & 0 & 1 & 2 & 0 & 0 & 0 & 0 & 6 & 870 & 0 & 12 & 8 & 0 & 0 & 0 & 197 & 59 & nsubj+xcomp \\
	\textbf{} & \textbf{nummod} & 0 & 0 & 0 & 3 & 3 & 0 & 0 & 1 & 0 & 0 & 41 & 0 & 169 & 0 & 1 & 0 & 0 & 7 & 12 & 7 & det \\
	\textbf{} & \textbf{obj} & 2 & 1 & 2 & 0 & 1 & 1 & 1 & 1 & 0 & 0 & 37 & 8 & 0 & 520 & 55 & 0 & 7 & 21 & 63 & 13 & obj+nmod \\
	\textbf{} & \textbf{obl} & 2 & 4 & 22 & 3 & 0 & 1 & 0 & 1 & 0 & 0 & 23 & 9 & 0 & 57 & 640 & 0 & 7 & 7 & 195 & 27 & obl+nmod \\
	\textbf{} & \textbf{parataxis} & 1 & 0 & 1 & 0 & 0 & 4 & 0 & 9 & 0 & 0 & 0 & 0 & 0 & 0 & 0 & 37 & 0 & 0 & 23 & 2 & obl+parataxis \\
	\textbf{} & \textbf{xcomp} & 0 & 6 & 6 & 0 & 1 & 3 & 0 & 0 & 0 & 0 & 1 & 1 & 0 & 6 & 8 & 0 & 141 & 7 & 18 & 3 & obj+amod \\
	\textbf{En\textbackslash Zh} & \textbf{} &  &  &  &  &  &  &  &  &  &  &  &  &  &  &  &  &  &  &  &  &  \\
	\textbf{} & \textbf{acl} & 96 & 2 & 1 & 5 & 1 & 0 & 6 & 0 & 0 & 0 & 3 & 25 & 0 & 6 & 3 & 0 & 3 & 2 & 143 & 8 & obj+advcl \\
	\textbf{} & \textbf{advcl} & 1 & 44 & 1 & 0 & 0 & 4 & 0 & 0 & 0 & 1 & 0 & 0 & 0 & 1 & 3 & 0 & 34 & 0 & 116 & 14 & dep \\
	\textbf{} & \textbf{advmod} & 4 & 8 & 227 & 7 & 0 & 2 & 7 & 0 & 0 & 0 & 2 & 5 & 1 & 15 & 26 & 0 & 13 & 52 & 227 & 19 & advmod+obj \\
	\textbf{} & \textbf{amod} & 46 & 1 & 20 & 194 & 1 & 0 & 351 & 0 & 0 & 0 & 56 & 28 & 38 & 5 & 8 & 0 & 1 & 160 & 206 & 35 & compound+compound \\
	\textbf{} & \textbf{appos} & 1 & 0 & 0 & 0 & 53 & 0 & 9 & 2 & 0 & 2 & 5 & 2 & 0 & 0 & 0 & 0 & 1 & 3 & 29 & 2 & compound+compound \\
	\textbf{} & \textbf{ccomp} & 0 & 1 & 2 & 0 & 0 & 37 & 0 & 0 & 0 & 0 & 0 & 0 & 0 & 0 & 0 & 0 & 5 & 0 & 54 & 4 & ccomp+advcl \\
	\textbf{} & \textbf{compound} & 2 & 0 & 0 & 15 & 3 & 0 & 317 & 4 & 0 & 14 & 26 & 9 & 1 & 2 & 3 & 0 & 0 & 250 & 109 & 64 & compound+compound \\
	\textbf{} & \textbf{conj} & 2 & 51 & 0 & 0 & 0 & 1 & 7 & 225 & 0 & 1 & 1 & 0 & 0 & 0 & 0 & 0 & 0 & 10 & 200 & 23 & dep \\
	\textbf{} & \textbf{fixed} & 0 & 0 & 0 & 0 & 1 & 0 & 0 & 0 & 0 & 1 & 1 & 0 & 0 & 0 & 0 & 0 & 0 & 60 & 1 & 1 & appos+flat \\
	\textbf{} & \textbf{flat} & 0 & 0 & 0 & 0 & 18 & 0 & 20 & 2 & 0 & 99 & 0 & 0 & 0 & 0 & 0 & 0 & 0 & 19 & 53 & 39 & appos+flat \\
	\textbf{} & \textbf{nmod} & 18 & 0 & 4 & 13 & 6 & 1 & 200 & 4 & 0 & 3 & 259 & 38 & 13 & 44 & 37 & 0 & 0 & 31 & 427 & 29 & compound+compound \\
	\textbf{} & \textbf{nsubj} & 2 & 2 & 5 & 0 & 0 & 1 & 5 & 0 & 0 & 0 & 5 & 674 & 0 & 31 & 14 & 0 & 1 & 1 & 391 & 92 & nsubj+advcl \\
	\textbf{} & \textbf{nummod} & 0 & 0 & 0 & 1 & 0 & 0 & 0 & 0 & 0 & 1 & 0 & 0 & 74 & 0 & 0 & 0 & 0 & 39 & 57 & 41 & nummod+clf \\
	\textbf{} & \textbf{obj} & 4 & 4 & 1 & 0 & 1 & 7 & 6 & 0 & 0 & 0 & 3 & 6 & 2 & 396 & 29 & 0 & 7 & 26 & 180 & 50 & ccomp+nsubj \\
	\textbf{} & \textbf{obl} & 4 & 6 & 21 & 0 & 0 & 13 & 8 & 0 & 0 & 0 & 8 & 34 & 0 & 138 & 159 & 0 & 18 & 14 & 425 & 49 & obj+advcl \\
	\textbf{} & \textbf{parataxis} & 0 & 2 & 1 & 0 & 0 & 6 & 0 & 0 & 0 & 0 & 0 & 0 & 0 & 0 & 0 & 0 & 1 & 0 & 71 & 18 & dep \\
	\textbf{} & \textbf{xcomp} & 1 & 5 & 11 & 0 & 1 & 24 & 0 & 0 & 0 & 0 & 0 & 1 & 0 & 9 & 2 & 0 & 42 & 5 & 62 & 22 & aux \\
	\textbf{En\textbackslash Ko} & \textbf{} &  &  &  &  &  &  &  &  &  &  &  &  &  &  &  &  &  &  &  &  &  \\
	\textbf{} & \textbf{acl} & 73 & 4 & 3 & 0 & 0 & 0 & 4 & 0 & 0 & 0 & 2 & 2 & 0 & 4 & 0 & 0 & 0 & 0 & 106 & 7 & acl+nsubj+acl \\
	\textbf{} & \textbf{advcl} & 2 & 24 & 9 & 0 & 0 & 0 & 0 & 0 & 0 & 1 & 0 & 2 & 0 & 0 & 0 & 0 & 0 & 0 & 95 & 8 & advmod+acl \\
	\textbf{} & \textbf{advmod} & 5 & 9 & 169 & 4 & 0 & 0 & 13 & 0 & 1 & 0 & 1 & 5 & 1 & 4 & 0 & 0 & 0 & 14 & 168 & 30 & aux \\
	\textbf{} & \textbf{amod} & 68 & 1 & 20 & 106 & 0 & 0 & 339 & 0 & 0 & 5 & 69 & 16 & 16 & 3 & 0 & 0 & 0 & 87 & 168 & 41 & det \\
	\textbf{} & \textbf{appos} & 25 & 1 & 0 & 0 & 7 & 0 & 16 & 0 & 0 & 3 & 2 & 2 & 0 & 0 & 0 & 0 & 0 & 0 & 37 & 4 & acl+compound \\
	\textbf{} & \textbf{ccomp} & 0 & 18 & 2 & 0 & 0 & 1 & 0 & 0 & 0 & 1 & 0 & 0 & 0 & 0 & 0 & 0 & 0 & 0 & 39 & 5 & advcl+advcl \\
	\textbf{} & \textbf{compound} & 2 & 1 & 4 & 0 & 0 & 0 & 354 & 7 & 0 & 21 & 14 & 4 & 2 & 4 & 0 & 0 & 0 & 143 & 67 & 13 & compound+compound \\
	\textbf{} & \textbf{conj} & 2 & 47 & 2 & 0 & 0 & 0 & 2 & 201 & 0 & 3 & 2 & 0 & 0 & 1 & 0 & 0 & 0 & 2 & 118 & 7 & conj+compound \\
	\textbf{} & \textbf{fixed} & 0 & 0 & 0 & 0 & 0 & 0 & 2 & 0 & 0 & 0 & 0 & 0 & 0 & 0 & 0 & 0 & 0 & 3 & 2 & 1 & compound+compound \\
	\textbf{} & \textbf{flat} & 6 & 0 & 0 & 0 & 0 & 0 & 28 & 0 & 0 & 101 & 1 & 1 & 0 & 0 & 0 & 0 & 0 & 9 & 22 & 5 & compound+flat \\
	\textbf{} & \textbf{nmod} & 10 & 1 & 26 & 3 & 5 & 0 & 139 & 4 & 0 & 3 & 207 & 21 & 6 & 12 & 4 & 0 & 0 & 8 & 366 & 51 & acl+advmod \\
	\textbf{} & \textbf{nsubj} & 5 & 1 & 18 & 1 & 0 & 0 & 7 & 0 & 0 & 0 & 2 & 374 & 0 & 23 & 0 & 0 & 0 & 1 & 358 & 55 & nsubj+advcl \\
	\textbf{} & \textbf{nummod} & 0 & 0 & 2 & 0 & 0 & 0 & 3 & 0 & 0 & 0 & 0 & 0 & 100 & 0 & 0 & 0 & 0 & 1 & 40 & 14 & nummod+nmod \\
	\textbf{} & \textbf{obj} & 5 & 5 & 30 & 0 & 0 & 0 & 16 & 0 & 0 & 0 & 2 & 21 & 0 & 249 & 0 & 0 & 0 & 11 & 143 & 27 & aux+obj \\
	\textbf{} & \textbf{obl} & 5 & 7 & 210 & 0 & 0 & 0 & 12 & 0 & 0 & 0 & 2 & 24 & 0 & 35 & 2 & 0 & 0 & 4 & 350 & 25 & advmod+compound \\
	\textbf{} & \textbf{parataxis} & 1 & 9 & 0 & 0 & 0 & 0 & 0 & 0 & 0 & 0 & 0 & 0 & 0 & 0 & 0 & 0 & 0 & 0 & 33 & 5 & advcl+advcl \\
	\textbf{} & \textbf{xcomp} & 3 & 12 & 17 & 0 & 0 & 9 & 9 & 0 & 0 & 0 & 0 & 1 & 0 & 7 & 0 & 0 & 0 & 2 & 61 & 11 & aux \\
	\textbf{En\textbackslash Jp} & \textbf{} &  &  &  &  &  &  &  &  &  &  &  &  &  &  &  &  &  &  &  &  &  \\
	\textbf{} & \textbf{acl} & 42 & 5 & 0 & 0 & 0 & 0 & 1 & 0 & 0 & 0 & 9 & 1 & 0 & 1 & 0 & 0 & 0 & 5 & 58 & 6 & nmod+acl \\
    \textbf{} & \textbf{advcl} & 0 & 4 & 0 & 0 & 0 & 0 & 1 & 0 & 0 & 0 & 0 & 0 & 0 & 1 & 0 & 0 & 0 & 4 & 15 & 2 & obj+acl \\
    \textbf{} & \textbf{advmod} & 2 & 6 & 37 & 3 & 0 & 1 & 10 & 0 & 0 & 0 & 3 & 4 & 6 & 2 & 5 & 0 & 0 & 31 & 112 & 21 & aux \\
    \textbf{} & \textbf{amod} & 73 & 2 & 3 & 63 & 0 & 0 & 185 & 0 & 0 & 0 & 131 & 5 & 15 & 5 & 1 & 0 & 0 & 222 & 108 & 17 & compound+compound \\
    \textbf{} & \textbf{appos} & 4 & 2 & 0 & 0 & 11 & 0 & 21 & 0 & 0 & 0 & 25 & 0 & 0 & 0 & 0 & 0 & 0 & 2 & 41 & 13 & nmod+nmod \\
    \textbf{} & \textbf{ccomp} & 1 & 1 & 0 & 0 & 0 & 2 & 0 & 0 & 0 & 0 & 0 & 0 & 0 & 0 & 1 & 0 & 0 & 0 & 9 & 1 & aux \\
    \textbf{} & \textbf{compound} & 3 & 0 & 0 & 4 & 0 & 0 & 224 & 0 & 0 & 0 & 108 & 0 & 1 & 2 & 1 & 0 & 0 & 282 & 68 & 21 & compound+compound \\
    \textbf{} & \textbf{conj} & 2 & 18 & 0 & 2 & 0 & 0 & 1 & 0 & 0 & 0 & 121 & 1 & 0 & 0 & 2 & 0 & 0 & 7 & 188 & 51 & nmod+nmod \\
    \textbf{} & \textbf{fixed} & 0 & 0 & 0 & 0 & 0 & 0 & 1 & 0 & 0 & 0 & 1 & 0 & 0 & 0 & 0 & 0 & 0 & 44 & 2 & 1 & case \\
    \textbf{} & \textbf{flat} & 0 & 0 & 0 & 0 & 0 & 0 & 32 & 0 & 0 & 0 & 93 & 0 & 0 & 0 & 0 & 0 & 0 & 40 & 45 & 17 & nmod+compound \\
    \textbf{} & \textbf{nmod} & 2 & 1 & 0 & 2 & 2 & 0 & 72 & 0 & 0 & 0 & 378 & 3 & 1 & 6 & 7 & 0 & 0 & 68 & 329 & 63 & nmod+nmod \\
    \textbf{} & \textbf{nsubj} & 2 & 0 & 1 & 0 & 0 & 0 & 3 & 0 & 0 & 0 & 11 & 199 & 0 & 5 & 11 & 0 & 0 & 25 & 184 & 25 & nsubj+advcl \\
    \textbf{} & \textbf{nummod} & 1 & 0 & 0 & 0 & 0 & 0 & 5 & 0 & 0 & 0 & 0 & 0 & 68 & 0 & 1 & 0 & 0 & 11 & 28 & 17 & nummod+compound \\
    \textbf{} & \textbf{obj} & 0 & 3 & 0 & 1 & 0 & 0 & 8 & 0 & 0 & 0 & 12 & 17 & 0 & 93 & 3 & 0 & 0 & 22 & 44 & 10 & iobj \\
    \textbf{} & \textbf{obl} & 1 & 1 & 9 & 0 & 0 & 0 & 7 & 0 & 0 & 0 & 16 & 4 & 0 & 13 & 73 & 0 & 0 & 12 & 216 & 59 & iobj \\
    \textbf{} & \textbf{parataxis} & 0 & 1 & 0 & 0 & 0 & 1 & 0 & 0 & 0 & 0 & 0 & 0 & 0 & 0 & 0 & 0 & 0 & 0 & 12 & 2 & advcl+obl \\
    \textbf{} & \textbf{xcomp} & 3 & 5 & 2 & 0 & 0 & 3 & 0 & 0 & 0 & 0 & 0 & 1 & 0 & 2 & 1 & 0 & 0 & 2 & 20 & 7 & aux
\end{tabular}
}
\caption{Raw counts of corresponding syntactic relations of frequent UD edge types.}
\label{table:ud-divergences-edges-raw}
\end{table*}

\section{Translation entropies of UD relations}\label{ssec:rel-entropies}

Translation entropies of UD relations are shown in Table~\ref{tab:translation-entropies}.

\section{Zero-shot Parsing: Implementation Details}\label{sec:zero-shot-parsing-exp-setup}

We used the AllenNLP \citep{gardner-etal-2018-allennlp} implementation of the deep biaffine attention model by \citet{dozat2017biaffine}. The only modification is that we replaced the trainable Glove embeddings with multilingual Bert \citep{devlin2018bert} untrainable embeddings (i.e., we didn't perform any fine-tuning on Bert), using the built-in embeddings in AllenNLP with the default settings. 
We ignore UD sub-categorization of the edge labels (as the sub-types are often language-specific). The full list of hyper-parameters is given in Table \ref{tab:hyperparams}.

We trained three models for each language (with the same hyperparameters), using the UD v2.5 English-EWT dataset for the English model, and the GSD datasets (also v2.5) for French, Russian, Chinese, Korean and Japanese. Standard splits were used for all corpora.\footnote{All UD corpora can be found in \url{https://github.com/UniversalDependencies/}}
The models were evaluated on the GSD and PUD datasets.

The per-label F-scores used for linear modelling in the paper are averages of the F-scores of the three models. The following relations were considered: \texttt{acl}, \texttt{advcl}, \texttt{advmod}, \texttt{amod}, \texttt{appos}, \texttt{ccomp}, \texttt{compound}, \texttt{conj}, \texttt{fixed}, \texttt{flat}, \texttt{nmod}, \texttt{nsubj}, \texttt{nummod}, \texttt{obj}, \texttt{obl}, \texttt{parataxis}, \texttt{xcomp}.\footnote{The list is slightly longer than that used in the automation experiments since thanks to the precision of manual alignment we were able to target relatively rare edge labels in a small corpus.} Not all of them were present in parser outputs for all languages due to training-set peculiarities, and missing relations were automatically omitted from the model. The R code used for fitting the models is available in the Supplementary Material.

\begin{table*}[ht]
	\centering
	\begin{tabular}{llllll}
		& \textbf{Ko} & \textbf{Jp} & \textbf{Zh} & \textbf{Ru} & \textbf{Fr} \\
	\textbf{acl} & 4.769242 & 3.830784 & 5.060668 & 3.652101 & 4.050027 \\
	\textbf{advcl} & 5.310528 & 2.846439 & 4.835997 & 4.504237 & 3.745798 \\
	\textbf{advmod} & 4.089253 & 5.092975 & 4.555394 & 2.848869 & 2.767038 \\
	\textbf{amod} & 3.551876 & 3.268781 & 3.642666 & 1.634234 & 1.709718 \\
	\textbf{appos} & 3.880474 & 3.21408 & 3.409938 & 3.297443 & 2.135566 \\
	\textbf{aux} & 3.087929 & 2.565961 & 2.013037 & 1.157876 & 0.508315 \\
	\textbf{case} & 3.50321 & 1.988427 & 3.81068 & 2.219851 & 1.333314 \\
	\textbf{cc} & 3.062907 & 4.349407 & 3.109744 & 1.773011 & 1.736929 \\
	\textbf{ccomp} & 4.411585 & 3.277613 & 4.158406 & 2.994698 & 2.65771 \\
	\textbf{compound} & 2.309827 & 2.145895 & 2.503707 & 2.988854 & 2.821717 \\
	\textbf{conj} & 3.531517 & 4.126376 & 4.16811 & 2.144241 & 2.039554 \\
	\textbf{cop} & 2.272808 & 3.21288 & 2.217022 & 3.000973 & 1.584963 \\
	\textbf{csubj} & 3.251629 & 0.811278 & 3.386637 & 3.221252 & 2.879249 \\
	\textbf{det} & 3.107192 & 2.769522 & 2.989865 & 2.689149 & 2.197421 \\
	\textbf{fixed} & 1.842371 & 0 & 0.46229 & 1.35203 & 2.584963 \\
	\textbf{flat} & 2.128367 & 2.381739 & 2.39878 & 1.634478 & 1.932195 \\
	\textbf{iobj} & 2.321928 & NA & 2.5 & 1.5 & 1.351644 \\
	\textbf{mark} & 2.96381 & 3.027169 & 4.024999 & 2.7912 & 2.142446 \\
	\textbf{nmod} & 4.971409 & 3.542603 & 5.006165 & 3.155123 & 2.795594 \\
	\textbf{nsubj} & 4.11521 & 4.069544 & 3.196435 & 2.319142 & 1.721087 \\
	\textbf{nummod} & 1.967499 & 1.946163 & 2.15335 & 1.120919 & 1.478528 \\
	\textbf{obj} & 3.488357 & 3.077944 & 3.095776 & 2.679303 & 1.934877 \\
	\textbf{obl} & 5.265461 & 4.585266 & 5.601495 & 3.664774 & 2.746977 \\
	\textbf{parataxis} & 4.446289 & 3.321928 & 4.897466 & 2.711151 & 3.091764 \\
	\textbf{xcomp} & 4.609696 & 3.039149 & 3.912925 & 3.054541 & 1.997426 \\
	\textbf{Average} & 3.53 & 3.02 & 3.48 & 2.56 & 2.24
	\end{tabular}
	\caption{Translation entropies of UD relations.}\label{tab:translation-entropies}
\end{table*}

\begin{table*}[t]
\centering
\begin{tabular}{|l|l|}

\hline
\multirow{2}{*}{Input} & Input dropout rate: 0.3 \\
& POS tag embedding dimension: 100 \\
\hline
\multirow{5}{*}{Word-level BiLSTM} & LSTM size: 400 \\
& \# LSTM layers: 3 \\
& Recurrent dropout rate: 0.3 \\
& Use Highway Connection: Yes \\
& Output dropout rate: 0.3 \\
\hline
\multirow{5}{*}{MLP and Attention} & Arc MLP size: 500 \\
& Label MLP size: 100 \\
& \# MLP layers: 1 \\
& Activation: ReLU \\
& Dropout: 0.3 \\
\hline
\multirow{6}{*}{MLP and Attention} &
Batch size: 128 \\
& \# Epochs: 100 \\
& Early stopping: 50 \\
& Adam \citep{kingma2014adam} lrate: 0.001 \\
& Adam $\beta1$: 0.9 \\
& Adam $\beta2$: 0.9 \\
\hline
\end{tabular}
\caption{Hyper-parameters used in our zero-shot experiments.\label{tab:hyperparams}}
\end{table*}


 \bibliography{propose}
 \bibliographystyle{acl_natbib}